\documentclass[10pt,twocolumn,letterpaper]{article}

\usepackage[pagenumbers]{cvpr} 

\usepackage[dvipsnames]{xcolor}

\usepackage{graphicx}
\usepackage{booktabs}
\usepackage[dvipsnames]{xcolor}
\usepackage{algorithm,algpseudocode}
\usepackage{amsmath, amssymb, amscd, mathrsfs,bm,tabularx}
\usepackage{enumitem}
\usepackage{colortbl}
\newsavebox{\measurebox}
\usepackage{multirow}

\usepackage[marginal]{footmisc}
\renewcommand{\thefootnote}{\fnsymbol{footnote}}
\usepackage[page]{appendix} 

\usepackage{titletoc}
\usepackage[accsupp]{axessibility}

\definecolor{cvprblue}{rgb}{0.21,0.49,0.74}
\usepackage[pagebackref,breaklinks,colorlinks,citecolor=cvprblue]{hyperref}

\def\paperID{33} 
\def\confName{3DV\xspace}
\def\confYear{2025\xspace}

\title{SAMPro3D: Locating SAM Prompts in 3D for Zero-Shot Instance Segmentation}

\author{
Mutian Xu\textsuperscript{\rm 1} \quad Xingyilang Yin\textsuperscript{\rm 1} \quad Lingteng Qiu\textsuperscript{\rm 1} \quad Yang Liu\textsuperscript{\rm 3} \quad Xin Tong\textsuperscript{\rm 3} \quad Xiaoguang Han\textsuperscript{\rm 1,2}\thanks{Corresponding author} \vspace{5pt}\\
{\normalsize \textsuperscript{\rm 1}{SSE, CUHKSZ} \qquad \textsuperscript{\rm 2}{FNii, CUHKSZ} \qquad \textsuperscript{\rm 3}{Microsoft Research Asia}} \vspace{5pt}\\
\small{\href{https://mutianxu.github.io/sampro3d/}{mutianxu.github.io/sampro3d}}
}

\begin{document}
\maketitle


\begin{abstract}
We introduce SAMPro3D for zero-shot instance segmentation of 3D scenes. Given the 3D point cloud and multiple posed RGB-D frames of 3D scenes, our approach segments 3D instances by applying the pretrained Segment Anything Model (SAM) to 2D frames. Our key idea involves locating SAM prompts in 3D to align their projected pixel prompts across frames, ensuring the view consistency of SAM-predicted masks.
Moreover, we suggest selecting prompts from the initial set guided by the information of SAM-predicted masks across all views, which enhances the overall performance.
We further propose to consolidate different prompts if they are segmenting different surface parts of the same 3D instance, bringing a more comprehensive segmentation.
Notably, our method does \textbf{not} require any additional training. Extensive experiments on diverse benchmarks show that our method achieves comparable or better performance compared to previous zero-shot or fully supervised approaches, and in many cases surpasses human annotations.
Furthermore, since our fine-grained predictions often lack annotations in available datasets, we present ScanNet200-Fine50 test data which provides fine-grained annotations on 50 scenes from ScanNet200 dataset.
\end{abstract}    
\section{Introduction}
\label{sec:intro}

Instance segmentation of 3D scenes plays a vital role in diverse applications such as augmented reality, room navigation, and autonomous driving. The objective is to predict 3D instance masks from input 3D scenes which are often represented by meshes, point clouds, and posed RGB-D images.
Traditional methods for this task \cite{pointgroup,minkowski,PointContrast,csc,kpconv,paconv,pointtrans,mask3d} lack the \textit{zero-shot} capability. They often struggle to accurately segment newly introduced object categories that were not encountered during training \cite{MSeg_2020_CVPR,michele2021generative}. 
Despite recent efforts \cite{openscene,pla,chen2023clip2scene,liu2023segment,sa3d,openmask3d,open3dis,openins3d,lowis3d,segment3D,samgraphcut,dong2023leveraginglargescalepretrainedvision} that harness vision foundation models \cite{radford2021clip,caron2021dino,kirillov2023sam} to enhance zero-shot 3D scene segmentation, they necessitate either 3D pretrained networks or training on domain-specific data. As a result, directly applying them to novel 3D scenes remains challenging in terms of generalization.

\begin{figure}[t]
  \centering
  \captionsetup{type=figure}
   \includegraphics[width=0.5\linewidth]{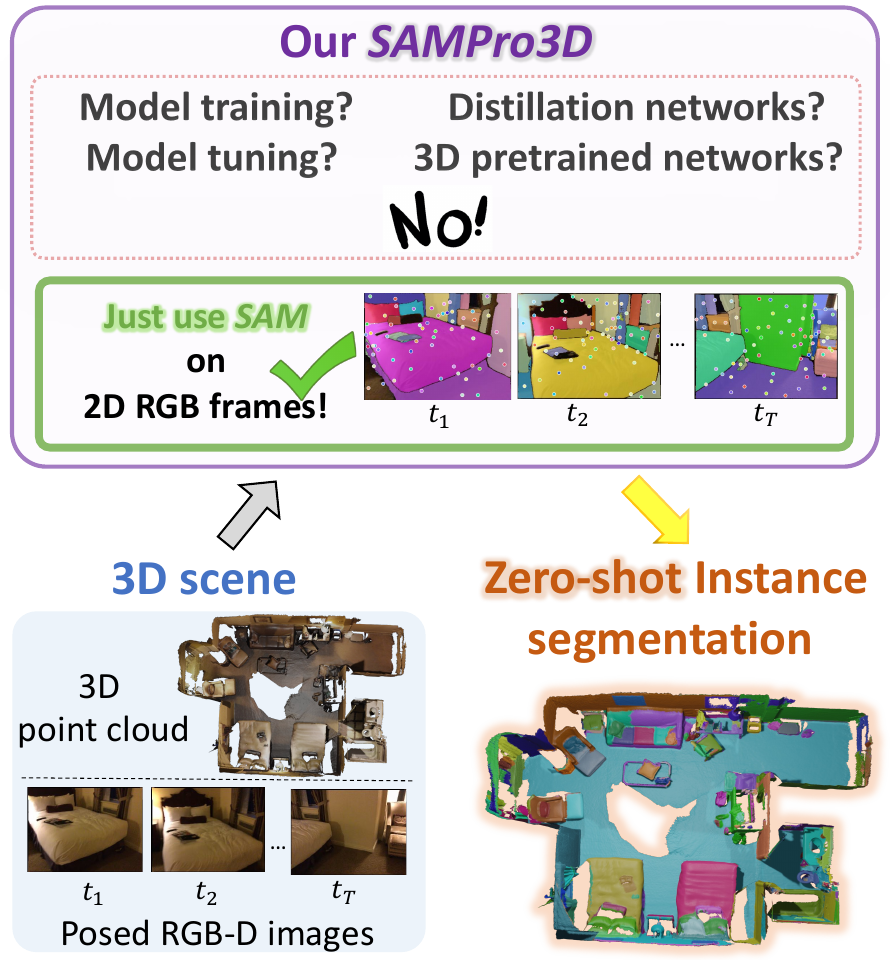}
   \vspace{-0.2cm}
   \captionof{figure}{We introduce \textit{SAMPro3D} for \textit{zero-shot} instance segmentation of 3D scenes.
   Given the 3D point cloud and posed RGB-D frames of 3D scenes, our approach uses the Segment Anything Model (SAM) \cite{kirillov2023sam} on RGB frames to segment 3D instances. Our method does \textbf{not} require additional training on domain-specific data. See \cref{fig:qualitative_result} for more impressive results.}
   \label{fig:teaser}
   \vspace{-0.5cm}
\end{figure}

In the field of 2D image segmentation, the Segment Anything Model (SAM) \cite{kirillov2023sam} brought the breakthrough.
Trained on an extensive SA-1B dataset \cite{kirillov2023sam}, SAM can “segment any unfamiliar images” without further training, by accepting various input prompts that specify where or what is to be segmented in an image.
The final stage of SAM (referred to as automatic-SAM) automatically generates prompts and corresponding segmentations on the image, where each single prompt accurately segments one 2D instance.
Having witnessed the great power of SAM, and recognizing that a 3D scene is essentially a combination of multiple 2D views, a curious \textbf{question} arises: Given 3D scene point clouds with posed RGB-D frames,
is it possible to apply SAM \textit{directly} to 2D frames for zero-shot 3D instance segmentation \textit{without} additional training?

\begin{figure*}[ht]
  \centering
  \captionsetup{type=figure}
   \includegraphics[width=0.65\linewidth]{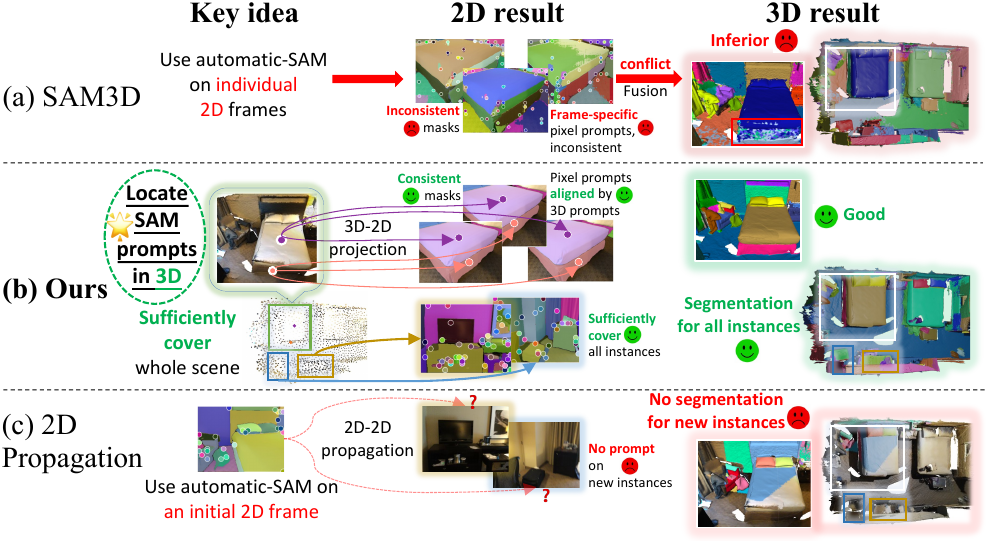}
   \vspace{-0.2cm}
   \captionof{figure}{\textbf{The comparison of our key idea and others}. Our method (b) \textit{locates SAM prompts in \textbf{3D}}, which aligns pixel prompts across frames, bringing the frame consistency of prompts and their masks, and can handle newly emerged instances.
   Here we use random colors to visualize 3D results for instance discrimination, so there is no correlation between the colors assigned to 2D and 3D instances.}
   \label{fig:key_idea}
   \vspace{-0.4cm}
\end{figure*}


There are several potential attempts to explore this question: (1) A recent project called SAM3D \cite{sam3d} utilizes automatic-SAM to individual 2D frames.
The resulting 2D masks are then projected into 3D space and iteratively fused to obtain the final result.
However, SAM3D assigns \textit{frame-specific} pixel prompts that lack view consistency, producing inconsistent 2D masks across frames.
As a result, fusing these masks may cause substantial conflicts in the segmentation of the same area, yielding inferior 3D results (see \cref{fig:key_idea} (a)).
(2) To realize prompt consistency across frames, one possible solution is to employ automatic-SAM on an initial frame to generate 2D-pixel prompts, which can then be propagated to subsequent frames, analogous to SAM-PT \cite{sam-pt} for video tracking.
Nevertheless, while videos processed by SAM-PT typically involve foreground objects that consistently appear across all frames, 3D scenes pose a different challenge. In 3D scenes, instances that initially appear may disappear in subsequent frames, and new instances may emerge. Consequently, initial prompts cannot be propagated to cover \textit{newly emerged} instances in other frames of 3D scenes, resulting in the absence of segmentation masks for these instances. (see \cref{fig:key_idea} (c)). 

To tackle the aforementioned challenges, we present a simple yet effective method called SAMPro3D. The key idea of SAMPro3D is to \textit{locate SAM prompts in \textbf{3D}} scene point clouds and project these 3D prompts onto 2D frames to get pixel prompts for using SAM (see \cref{fig:key_idea} (b)). In this way, a 3D input point serves as the natural prompt to align the pixel prompts projected from this 3D point across different frames, making both pixel prompts and their SAM-predicted masks for the same 3D instance exhibit consistency across frames.
Moreover, as our 3D prompts sufficiently cover all instances, we can obtain 3D segmentations for all instances in the scene.

Building upon this key idea, our framework is designed in a bottom-up manner. It begins by sampling 3D points from the input scene as the initial SAM prompts. Subsequently, we introduce a novel View-Guided Prompt Selection algorithm to select these initial prompts. It examines the quality of the segmentation masks generated by initial prompts within each view and accumulates the examinations across \textit{all views}. By doing so, it selects and retains high-quality prompts for segmenting each instance, thereby enhancing the overall performance. 
However, we find that in some cases, a single retained prompt may only segment part of a 3D instance due to the limited visible field of 2D camera views. 
To address this, we further design a Surface-Based Prompt Consolidation strategy to consolidate 3D prompts that exhibit certain intersections in their 3D masked \textit{surfaces} into one single prompt, as they are likely segmenting different parts of the same 3D instance. This brings a more comprehensive segmentation of 3D instances.
Finally, we project all input points of the scene onto each segmented frame and accumulate their predictions across frames to derive the final 3D segmentation.

Our underlying design logic is to build an automatic-SAM tailored for 3D instance segmentation. We aim to \textit{automatically} generate SAM prompts and ensure the consistency of their 2D segmentation for the same 3D instance across different frames, ultimately achieving that every \textit{single} prompt accurately segments \textit{one} 3D instance.
Notably, our method does \textbf{not} require additional training or 3D pretrained networks on domain-specific data. This preserves SAM's zero-shot ability and enables future applications to directly segment new 3D scenes without the need to gather plenty of training data.

We conduct extensive experiments on diverse benchmarks, including ScanNet \cite{scannet} containing indoor scenes, ScanNet200 \cite{scannet200} providing more comprehensive annotations, ScanNet++ \cite{scannetpp} offering more detailed segmentation masks, and KITTI-360 \cite{kitti360} featuring outdoor suburban scenes, where our approach demonstrates high effectiveness. In addition, we observed that our method often segments fine-grained instances that lack annotations in available datasets. For better evaluation, we present \textit{ScanNet200-Fine50} test data, providing more fine-grained annotations of 50 scenes from ScanNet200 validation set.

Our contributions are summarized as:
\begin{itemize}[noitemsep,topsep=0pt]
	\item[$\bullet$] To the best of our knowledge, we are the \textbf{first} to set SAM prompts on \textit{3D} surfaces and \textit{scatter} prompts to 2D views for 3D segmentation. This ensures the identity consistency of the 2D prompts across frames and can deal with emerging instances in other frames.
        \item[$\bullet$] Based on this key idea, we propose SAMPro3D for zero-shot 3D instance segmentation, equipped with novel prompt selection and consolidation that effectively enhance segmentation quality and comprehensiveness.
	\item[$\bullet$] Rich experiments show that our method consistently achieves higher quality and more diverse segmentations than previous zero-shot or fully supervised approaches, and in many cases surpasses human-level annotations.
        \item[$\bullet$] We present ScanNet200-Fine50 test data with more fine-grained annotations.
\end{itemize}

\section{Related Work}
\label{sec:related_work}

\paragraph{Closed-set 3D scene understanding.}
The field of 3D scene understanding has been dominated by closed set methods which primarily focus on training deep neural networks on domain-specific datasets \cite{sunrgbd,kitti,s3dis,scenenn,scannet,matterport3D,waymo,nuscenes,toscene,openroom}. The first line of research focuses on improving representation learning from human-annotated 3D labels \cite{pointnet,pointnet++,dgcnn,spidercnn,pointcnn,wu2019pointconv,pointwisecnn,minkowski,fpconv,closerlook,paconv,ma2022rethinking,pointtrans}, for solving different 3D scene understanding tasks \cite{votenet,PointRCNN,chen2020scanrefer,3detr,groupfree,chen2023voxenext,sparseconv,pointgroup,pointweb,mask3d}.
Another stream of works aims to construct semi-/weak-/self- supervision signals from 3D data \cite{PointContrast,huang2021spatio,4dcontrast,xu2022mm3Dscene,huang2023ponder,zhao2020sess,XuLee_CVPR20,chibane2021box2mask,cheng2021sspc,liu2022weakly,unscene3d}, so as to minimize the need for 3D annotations.
In addition, some methods leverage 2D supervision to assist the model training for 3D scene understanding \cite{vineet2015incremental,mccormac2017semanticfusion,kundu2020virtual,genova2021learning,sautier2022image,wang2022detr3d}.

However, the aforementioned methods all rely on training with domain-specific 3D or 2D data, limiting their zero-shot ability to understand new scenes that have never been seen during training.
Instead, our framework seeks to straightforwardly harness the inherent zero-shot ability of SAM for segmenting 3D scenes, thereby eliminating the need for additional model training.

\vspace{-0.3cm}
\paragraph{Zero-shot and open-set 3D scene understanding.}
The early studies on zero-shot 3D scene understanding are very limited \cite{michele2021generative,yang2023zero} and they still involve training with supervised 3D labels.
In recent years, many 2D vision foundation models have shown their remarkable zero-shot recognition abilities \cite{radford2021clip,caron2021dino,oquab2023dinov2,wang2023segGPT,zou2023seem,zou2023xdecoder,kirillov2023sam}.
This encourages the researchers to leverage them for 3D scene parsing. For example, \cite{chen2023clip2scene,openscene,openmask3d,open3dis}
all use CLIP \cite{radford2021clip} to extract pixel-wise features and align them with 3D space to realize language-guided segmentation of 3D scenes. \cite{pla,lowis3d} utilize \cite{wang2022ofa} to caption multi-view images for associating 3D and open-vocabulary concepts. Liu \etal \cite{liu2023segment} distill knowledge from \cite{kirillov2023sam,zou2023seem,zou2023xdecoder} to apply self-supervised learning on outdoor scenes. UnScene3D \cite{unscene3d} lifts DINO \cite{caron2021dino} to initialize 3D features for self-training.
 
Despite recent advancements in open-set methods, applying them directly to new 3D data remains challenging, since they still necessitate model tuning \cite{unscene3d}, 3D-2D distillation \cite{chen2023clip2scene,openscene,pla,lowis3d,liu2023segment}, or 3D/2D pretrained region proposal network \cite{openmask3d,ovir3d,open3dis,openins3d}. 
In contrast, our method just uses SAM \cite{kirillov2023sam} on RGB frames without requiring any of the aforementioned factors. This enables direct deployment for segmenting novel 3D scenes.

\vspace{-0.3cm}
\paragraph{The Segment Anything model.}
The Segment Anything Model (SAM) \cite{kirillov2023sam} has brought a revolution for image segmentation. Trained on an astonishing SA-1B dataset, SAM can effectively segment unfamiliar images without further training.
The distinguishing characteristic of SAM lies in its promptability, allowing it to accept various input prompts, which specify where or what is to be segmented in an image.
Several recent works are striving to lift SAM into 3D visual tasks.
Cen \etal \cite{sa3d} use SAM to segment target objects in NeRF \cite{mildenhall2020nerf} via one-shot manual prompting. 
Zhang \etal \cite{zhang2023sam3d} define hand-crafted grid prompts on Bird’s Eye View images and perform SAM for 3D object detection. 
SAM3D \cite{sam3d} employs automatic-SAM on individual 2D frames to generate 2D segmentation masks, which are then projected into 3D and gradually fused to derive the final 3D segmentation. 
Similarly, SAI3D \cite{sai3d} over-segments a 3D scene into superpoints, which are then progressively merged according to SAM masks.
However, both SAM3D and SAI3D all assign frame-specific prompts that lack consistency, causing segmentation conflicts across frames, finally yielding subpar 3D segmentation results.
Very recently, SAM has been applied to generate training labels \cite{segment3D} or graph annotations \cite{samgraphcut} for 3D segmentation, yet these methods still require training on domain-specific data.

\begin{figure*}[ht]
    \centering
    \captionsetup{type=figure}
    \includegraphics[width=1.0\textwidth]{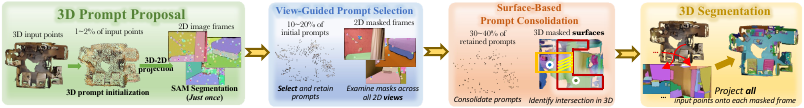}
    \captionof{figure}{\textbf{An overview of our \textit{SAMPro3D}}, with a primary focus on \textbf{\textit{“prompt”}}. Given 3D scene point clouds with posed RGB-D frames, we locate SAM \cite{kirillov2023sam} prompts in input 3D scenes and project them onto 2D frames to obtain 2D segmentation masks. Later, the initial prompts and their masks are selected (\cref{alg:prompt_filter}) and consolidated (\cref{fig:prompt_consolidation}), leveraging both multi-view and surface information. Finally, we project all input points onto each segmented frame to obtain the 3D segmentation result. 
    } \label{fig:pipeline}
    \vspace{-0.3cm}
\end{figure*}%
 
Different from them, our key idea is to locate SAM prompts in 3D space so that the pixel prompts derived from different frames but projected by the same 3D prompt become harmonized in the 3D space, leading to the frame consistency of prompts and their masks, and bringing high-quality 3D segmentation.


\section{Method}
\label{sec:method}

An overview of our framework is presented in \cref{fig:pipeline}, which is designed in a bottom-up manner.

\subsection{3D Prompt Proposal}\label{sec:prompt_proposal}

\paragraph{3D prompt initialization.}
Given a point cloud $\mathbf{F}\in \mathbb{R}^{N\times 3}$ of a 3D scene with $N$ points, we first employ furthest-point sampling (FPS) to sample $M$ points as the initial 3D prompt $\mathbf{P}\in \mathbb{R}^{M\times 3}$. FPS helps us to achieve a decent coverage of instances within a scene.
For simplification, we use $\mathbf{f} \in \mathbb{R}^{3}$ and $\mathbf{p} \in \mathbb{R}^{3}$ to denote an individual input point and a single 3D prompt, respectively. 
\vspace{-0.3cm}

\paragraph{3D-2D projection.}
Following \cite{openscene}, we only consider the pinhole camera configuration here. In particular, given the camera intrinsic matrix $I_i$ and world-to-camera extrinsic matrix $E_i$ of a frame $i$, we calculate the corresponding pixel projection $\mathbf{x}=(u, v)$ of a point prompt $\mathbf{p}$ by: 
$\tilde{\mathbf{x}} = I_i \;\cdot E_i \;\cdot \tilde{\mathbf{p}}$, where $\tilde{\mathbf{x}}$ and $\tilde{\mathbf{\mathbf{p}}}$ are the homogeneous coordinates of $\mathbf{x}$ and $\mathbf{p}$, respectively. Similar to \cite{openscene,openmask3d}, an occlusion test is performed using depth values, to ensure that the pixel $\mathbf{x}$ is only valid when its corresponding point $\mathbf{p}$ is visible in frame $i$.
\vspace{-0.3cm}

\paragraph{SAM segmentation on image frames.}
SAM \cite{kirillov2023sam} is a promptable segmentation model that can accept various inputs such as pixel coordinates or bounding boxes and predict the segmentation area associated with each prompt.
In our framework, we feed all pixel coordinates calculated before to prompt SAM and obtain the 2D segmentation masks on all frames.
As depicted in \cref{fig:key_idea} and described in \cref{sec:intro}, through locating prompts in 3D space, the pixel prompts originating from distinct frames but projected by the identical 3D prompt will be aligned in 3D space, bringing the frame consistency on pixel prompts and their SAM-predicted masks.
Notably, this step is the \textit{only} step that we need to perform SAM. 
In the later stages, our attention is directed toward selecting and consolidating initial 3D prompts along with their segmentation masks. 

\subsection{View-Guided Prompt Selection}\label{sec:prompt_filter}

\begin{figure}[t]
\vspace{-0.3cm}
\begin{algorithm}[H]
\footnotesize
\caption{- View-Guided Prompt Selection}\label{alg:prompt_filter}
\begin{algorithmic}[1]
\State $\mathbf{s}\gets 0$, $\mathbf{c}\gets 0$
\While{$i$ is a frame} \;  \# start single-view examination
  \If{$\mathbf{p}$ has a valid pixel projection $\mathbf{x}$ in current frame $i$}
   \State $\mathbf{c}\gets \mathbf{c}+1$
  \EndIf
  \State Perform prompt selection according to the information of their SAM-predicted masks
  \If{$\mathbf{x}$ is selected after this examination}
    \State $\mathbf{s}\gets \mathbf{s}+1$
  \EndIf
  \State $i\gets i+1$ \; \# go to the next frame
\EndWhile \;  \# finish examination on all frames
\State $\bm{\theta} = \mathbf{s} \;/\; \mathbf{c}$
\If{$\bm{\theta} > \mathtt{\theta}_{retain}$}
 \State Retain this 3D prompt $\mathbf{p}$
\EndIf
\end{algorithmic}
\end{algorithm}
\vspace{-0.8cm}
\end{figure}

After the prompt initialization, although consistent, multiple prompts may segment the same instance, causing redundancy in the segmentation. Besides, some prompts may generate inaccurate masks that hurt the performance.
To handle this, we introduce a View-Guided Prompt Selection algorithm to select prompts.

As outlined in \cref{alg:prompt_filter}, we examine SAM-predicted masks and accumulate the examinations across all views. First, in each view, if a 3D prompt $\mathbf{p}$ has a valid pixel projection $\mathbf{x}$, its counter $\mathbf{c}$ increments. Next, if its 2D mask does not have overlaps with other masks, we select it as necessary for segmenting an instance. If several 2D masks exhibit significant overlaps, we select the ones with the highest SAM confidence value as the most representative prompt. The score $\mathbf{s}$ of a selected prompt will be accumulated.
After examining all views, we compute the probability of retaining a 3D prompt by $\bm{\theta} = \mathbf{s} \;/\; \mathbf{c}$, and retain the prompt when its probability exceeds a predefined threshold $\mathtt{\theta}_{retain}$.

This algorithm enables us to utilize multi-view information from all 2D views. It prioritizes high-quality prompts while maintaining prompt consistency, ultimately enhancing the 3D segmentation result.
It is ablated in \cref{sec:ablation}. More details are provided in the supplementary material.

\subsection{Surface-Based Prompt Consolidation}\label{sec:prompt_consolidation}
\begin{figure}[t]
  \centering
   \includegraphics[width=0.56\linewidth]{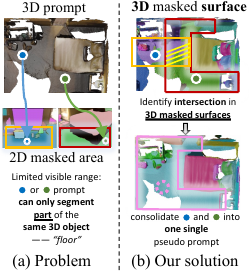}
   \caption{The illustration of the partial segmentation problem and our Surface-based Prompt Consolidation strategy.}
   \label{fig:prompt_consolidation}
   \vspace{-0.4cm}
\end{figure}

We have observed that in some cases, a single retained prompt may only segment part of a 3D instance due to the limited visible field of 2D camera views.
This issue is particularly prominent for large-sized instances that require multiple 2D views to be fully captured.
As in \cref{fig:prompt_consolidation} (a), several prompts segment different parts of the floor.

To address this, instead of solely using multi-view information, we further explicitly leverage 3D surface information and develop a Surface-Based Prompt Consolidation strategy (see \cref{fig:prompt_consolidation} (b)). This strategy involves checking the 3D masked \textit{surfaces} generated by different 3D prompts and identifying a certain intersection between them in 3D space. In such cases, we consider these prompts as likely segmenting the same 3D instance and consolidate them into a single pseudo prompt.
This process facilitates the integration of 3D information across prompts, leading to a more comprehensive segmentation of 3D instances.

The prompt consolidation performs stably well in our experiment. The key to consolidation lies in identifying the \textit{intersection} between surfaces segmented by prompts across views. Since 2D sequences have small view changes (even when skipping 20 frames as in supplementary material), there is always an intersection between continuous masked surfaces that segment the same 3D instance.

\subsection{3D Scene Segmentation}\label{sec:3d_segmentation}
After previous procedures, we have obtained the final set of 3D prompts and their 2D segmentation masks across frames. 
In addition, we have also ensured that each single prompt segment one 3D instance, allowing the \textit{prompt ID} to naturally serve as the \textit{instance ID}.
With the ultimate goal of segmenting all points within the 3D scene, we continue by projecting \textit{all} input points of the scene onto each segmented frame and compute their predictions using the following steps:
For each individual input point $\mathbf{f}$ in the scene, if it is projected within the mask area segmented by a prompt $p_k$ in frame $i$, we assign its prediction within that frame as the prompt ID $k$. 
We accumulate the predictions of $\mathbf{f}$ across all frames and assign its final prediction ID based on the prompt ID that has been assigned to it the most number of times.
By repeating this for all input points, we can achieve a complete 3D segmentation of the input scene.

In our study, all points can successfully get a valid mask, as it is \textit{easy to achieve}:
i) Simply initializing adequate prompts ensures comprehensive scene coverage with their generated masks, yet using only a few prompts may cause segmentation absence (\cref{fig:ablation}: 0.1\%). ii) Even if some frames have pixels without masks, there are sufficient frames to provide masks. As in supplementary material, all points can get a mask although skipping 20 frames. iii) If a point fails to get a mask (not observed in our experiment), we can assign it the label that occurs most frequently among its neighboring labeled points.
\section{Experiments}
\label{sec:experiments}

\paragraph{Setup.}
We conduct experiments on diverse 3D scene datasets, including ScanNet (v2) \cite{scannet}, ScanNet200 \cite{scannet200} and ScanNet++ \cite{scannetpp} for indoor rooms, and KITTI-360 \cite{kitti360} for outdoor environments.
ScanNet contains 1513 RGB-D indoor scans, with estimated camera parameters, surface reconstructions, textured meshes, and semantic annotations.
ScanNet200 provides \textit{more extensive annotations} for 200 categories based on ScanNet.
ScanNet++ offers 280 indoor scenes with \textit{more detailed segmentation masks} and high-resolution RGB images.
ScanNet200 and ScanNet++ present the conspicuously challenging scenario for zero-shot scene segmentation.
KITTI-360 is an \textit{outdoor} scene dataset with 300 suburban scenes, comprising 320k images and 100k laser scans.
We use their official validation split for evaluation.
In the supplementary material, we also evaluate the robustness of our method on Matterport3D \cite{matterport3D} dataset which exhibits \textit{large view changes}.

\begin{table*}[ht]
\centering
\resizebox{0.86\linewidth}{!}{
\begin{tabular}{l|ccc|ccc|ccc|ccc}
\toprule
& \multicolumn{3}{c|}{\textit{\textbf{ScanNet}}} & 
\multicolumn{3}{c|}{\textit{\textbf{ScanNet200}}} &
\multicolumn{3}{c|}{\textit{\textbf{ScanNet++}}} &
\multicolumn{3}{c}{\textit{\textbf{KITTI-360}}} \\
Model
 & $\mathbf{AP}$ & $\mathbf{AP_{50}}$ & $\mathbf{AP_{25}}$ 
 & $\mathbf{AP}$ & $\mathbf{AP_{50}}$ & $\mathbf{AP_{25}}$ 
 & $\mathbf{AP}$ & $\mathbf{AP_{50}}$ & $\mathbf{AP_{25}}$ 
 & $\mathbf{AP}$ & $\mathbf{AP_{50}}$ & $\mathbf{AP_{25}}$ \\
\midrule
\multicolumn{12}{l}{\textbf{training-\textit{dependent}}} & \\
UnScene3D \cite{unscene3d} [CVPR'24] & 15.9 & 32.2 & 58.5 & - & - & - & - & - & - & - & - & - \\
Segment3D \cite{segment3D} [ECCV'24] & - & - & - & - & - & - & 12.0 & 22.7 & 37.8 & - & - & - \\
SAM-graph \cite{samgraphcut} [ECCV'24] & - & - & - & 22.1 & 41.7 & 62.8 & 15.3 & 27.2 & 44.3 & 23.8 & \textbf{37.2} & 49.1\\
\midrule
\multicolumn{12}{l}{\textbf{training-\textit{free}}} & \\
HDBSCAN \cite{hdbscan} [ICDMW'17] & 1.6 & 5.5 & 32.1 & 2.9 & 8.2 & 33.1 & 4.3 & 10.6 & 32.3 & 9.3 & 18.9 & 39.6 \\
Felzenszwalb \cite{felzenszwalb2004efficient} [IJCV'04] & 5.3 & 12.6 & 36.9 & 4.8 & 9.8 & 27.5 & 8.8 & 16.9 & 36.1 & - & - & - \\
SAM3D~\cite{sam3d} [ICCVW'23] & 6.3 & 17.9 & 47.3 & 12.1 & 38.6 & 54.1 & 3.0 & 7.9 & 22.3 & 4.6 & 10.6 & 26.0 \\
SAI3D~\cite{sai3d} [CVPR'24] & \textbf{30.8} & \textbf{50.5} & \textbf{70.6} & - & - & - & 17.1 & 31.1 & 49.5 & - & - & - \\
\rowcolor{pink!12} \textbf{Ours} & 24.3 & 45.7 & 67.7 & \textbf{26.3} & \textbf{47.2} & \textbf{68.6} & \textbf{20.3} & \textbf{35.6} & \textbf{53.2} & \textbf{24.3} & 34.7 & \textbf{52.8} \\
\bottomrule
\end{tabular}
}
\vspace{-0.2cm}
\caption{\textbf{Quantitative} comparison with diverse \textbf{open-set} methods on ScanNet, ScanNet200, ScanNet++ and KITTI-360.}
\vspace{-0.2cm}
\label{tab:main_result}
\end{table*}

\begin{table}[t]
\centering
\setlength{\tabcolsep}{3pt}
\resizebox{0.8\linewidth}{!}{
\begin{tabular}{lllcccccc}
\toprule
&  & Training Data & $\mathbf{AP}$ & $\mathbf{AP_{50}}$ & $\mathbf{AP_{25}}$ \\
\midrule
\multirow{3}{*}{ScanNet200}
& Mask3D         & GT of ScanNet200        & \textbf{53.3} & \textbf{71.9} & \textbf{81.6} \\
& Mask3D         & GT of ScanNet           & 45.1 & 62.6 & 70.5 \\
& \textbf{Ours}  & -                       & 26.3 & 47.2 & 68.6 \\
\midrule
\midrule
\multirow{3}{*}{\textbf{ScanNet++}}
& Mask3D         & GT of ScanNet200        &  4.6 & 10.5 & 22.9 \\
& Mask3D         & GT of ScanNet           &  3.7 &  7.9 & 15.6 \\
& \textbf{Ours}  & -                       & \textbf{20.3} & \textbf{35.6} & \textbf{53.2} \\
\midrule
\multirow{3}{*}{\textbf{KITTI-360}}
& Mask3D         & GT of ScanNet200        &  0.2 &  0.9 &  7.0 \\
& Mask3D         & GT of ScanNet           &  0.3 &  1.0 &  8.0 \\
& \textbf{Ours}  & -                       & \textbf{24.3} & \textbf{34.7} & \textbf{52.8} \\
\bottomrule
\end{tabular}
}
\vspace{-0.2cm}
\caption{\textbf{Quantitative} comparison with \textbf{closed-set} method Mask3D \cite{mask3d}. While Mask3D outperforms ours on ScanNet200 when trained on ScanNet or ScanNet200, it cannot generalize well to ScanNet++ and KITTI-360.}
\label{tab:vsmask3d}
\vspace{-0.4cm}
\end{table}

To expedite processing, we resize each RGB image frame to a resolution of 240$\times$320, which has proven to be sufficient for our method to generate high-quality 3D segmentation results.
We employ the ViT-H SAM \cite{kirillov2023sam} model, which is the default public model for SAM. The entire framework is executed on a single NVIDIA A100 GPU.
As our method is \textit{class-agnostic}, we benchmark it against the class-agnostic baselines to ensure a fair comparison. The semantic labels are not considered in the evaluation, following \cite{unscene3d,samgraphcut}. Additionally, we follow \cite{samgraphcut} to exclude the predicted instances in unannotated regions for all methods, facilitating a fairer comparison.
\vspace{-0.3cm}

\paragraph{Methods in comparison.}
(1) \textbf{Open-set} methods, which do not learn from manually annotated 3D labels from a predefined set. These methods can be further split into two categories: i) \textbf{training-\textit{dependent}} methods which require additional model tuning \cite{unscene3d,segment3D,samgraphcut} on domain-specific data.
ii) \textbf{training-\textit{free}} methods without additional training, containing traditional unsupervised methods \cite{hdbscan,felzenszwalb2004efficient} and SAM-based methods \cite{sam3d,sai3d}. As highlighted previously, our method directly applies SAM to RGB-frames, which also eliminates the need for further training.

Note that SAM3D \cite{sam3d} and Segment3D \cite{segment3D} additionally report the scores incorporating the result from Felzenszwalb's algorithm \cite{felzenszwalb2004efficient} to refine their original results. Here, to fairly compare the performance of the algorithms themselves, we present their results without post-processing.

(2) \textbf{Closed-set} method: we also compare our method with Mask3D \cite{mask3d}, a state-of-the-art fully-supervised method trained on predefined annotations. We use its official pretrained models that are trained on ScanNet and ScanNet200 training sets.
Additionally, Mask3D does not treat floor and wall as instances, resulting in the absence of these two labels in its results.

\paragraph{Quantitative metrics.}
We follow the standard of instance segmentation task defined in \cite{scannet,fasterrcnn}, calculating mean Average Precision (mAP) at IoU thresholds of 50\%, 25\%, and averaged from 50\% to 95\% with a step size of 5\% (denoted by $\mathbf{AP_{50}}$, $\mathbf{AP_{25}}$ and $\mathbf{AP}$, respectively).

\subsection{Main Results}

\begin{figure}[t]
    \centering
    \captionsetup{type=figure}
    \includegraphics[width=1.0\linewidth]{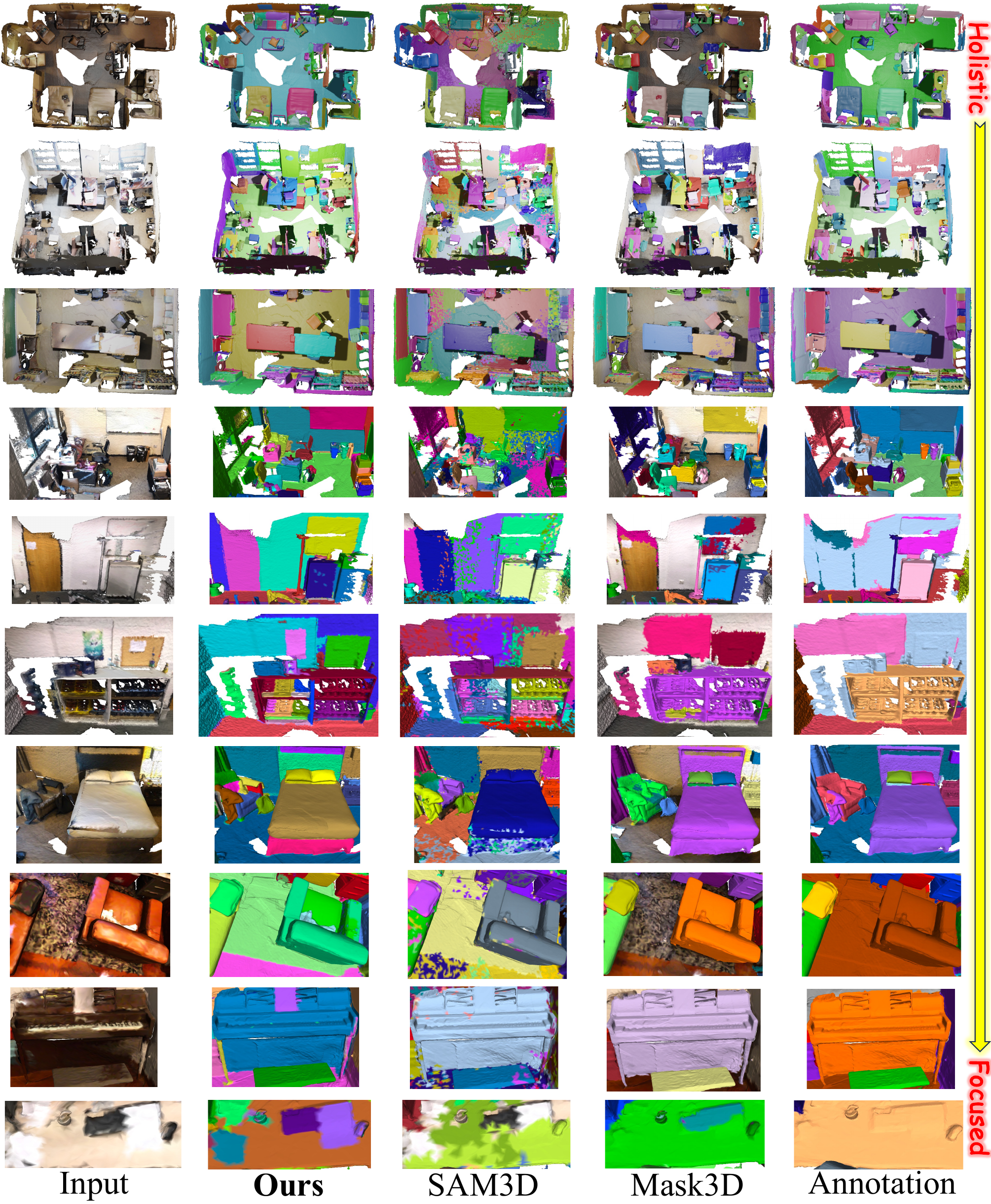}
    \captionof{figure}{\textbf{Qualitative comparison} of our method, SAM3D~\cite{sam3d}, Mask3D~\cite{mask3d} and ScanNet200's annotations \cite{scannet200}, across various scenes in ScanNet200, from holistic to focused view. Mask3D does not treat floor and wall as instances, resulting in the absence of these two labels in its results. Better view in zoom and color.
    }
    \label{fig:qualitative_result}
    \vspace{-0.5cm}
\end{figure}

\paragraph{Quantitative Results}\label{sec:quant_results}
(1) As in \cref{tab:main_result}, when compared with \textbf{open-set} methods, our training-free approach achieves comparable or better performance than training-dependent/free methods. While the recent method SAI3D [CVPR'24] surpasses ours on ScanNet, our approach excels on ScanNet++ (4.5$\uparrow$ $\mathbf{AP_{50}}$). To explain, ScanNet's annotations are generally coarse, whereas ScanNet++ offers detailed segmentation annotations for fine-grained instances. These detailed annotations hold more significance in evaluating \textit{zero-shot} capabilities, demonstrating our method's proficiency in segmenting \textit{fine-grained} instances, consistent with \cref{fig:qualitative_result}.
Moreover, our training-\textit{free} method outperforms the very recent training-\textit{dependent} method SAM-graph [ECCV'24] on all datasets, especially on ScanNet200 (5.5$\uparrow$ $\mathbf{AP_{50}}$) and ScanNet++ (8.4$\uparrow$ $\mathbf{AP_{50}}$).

(2) Following \cite{samgraphcut}, \cref{tab:vsmask3d} reports the comparison with the \textbf{closed-set} method, Mask3D.
When Mask3D is trained on ScanNet or ScanNet200 and then tested on ScanNet200, it outperforms our method.
However, its results significantly drop and lag behind ours when tested on ScanNet++. ScanNet++ is collected by the laser scanner, which is different from ScanNet gained by RGB-D fusion\cite{dai2017bundlefusion}. This shows that Mask3D is sensitive to data acquisition schemes.
Furthermore, when evaluated on the outdoor KITTI-360 which is distinct from indoor ScanNet200, Mask3D's performance becomes extremely poor.
This comparison highlights the zero-shot power of our approach.

\vspace{-0.2cm}
\paragraph{Qualitative results.}
\cref{fig:qualitative_result} shows qualitative comparisons on ScanNet200 \cite{scannet200}, across a variety of scenes (\eg, bedroom, office, classroom) and instances, from the holistic to the focused perspective.
Our method outperforms the open-set method SAM3D in terms of both segmentation quality and diversity.
When compared to the closed-set approach Mask3D (\textit{trained and evaluated both on ScanNet200}), our method achieves competitive or even superior segmentation accuracy and diversity. 
Moreover, compared to the extensive annotations of ScanNet200, our results are not only comparable in accuracy but also exhibit greater diversity in many cases.
More results a \textit{video demo} are provided in the supplementary material.

\begin{figure}[t]
\centering
    \centering
    \captionsetup{type=figure}
    \includegraphics[width=0.66\linewidth]{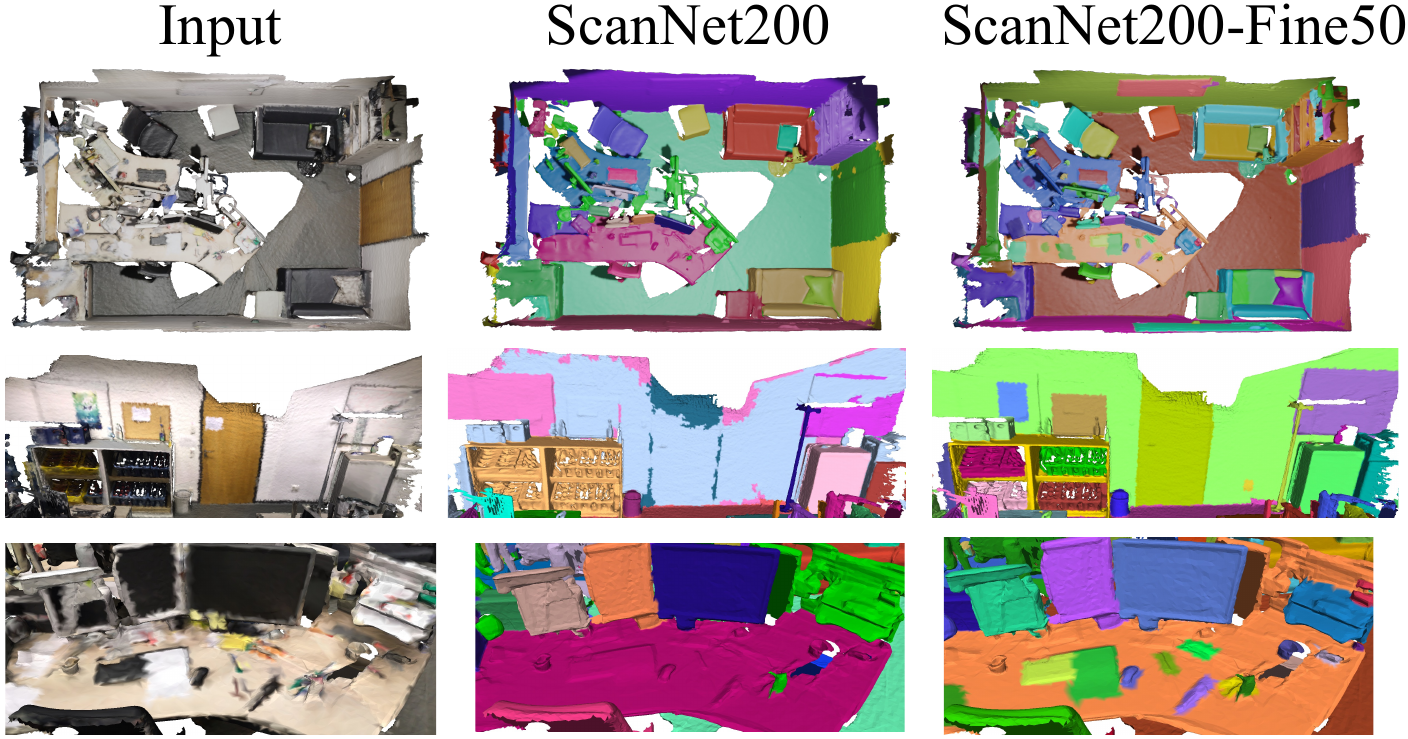}
    \vspace{-0.2cm}
    \captionof{figure}{Samples from ScanNet200 and our \textbf{ScanNet200-Fine50}.} 
    \label{fig:fine50}
    \vspace{-0.1cm}
\end{figure}

\begin{table}[t]
\centering
\resizebox{1.0\linewidth}{!}{
\begin{tabular}{lccc|ccc|ccc}
\toprule
Instance Size & \multicolumn{3}{c|}{\textbf{Normal}} & 
\multicolumn{3}{c|}{\textcolor{blue}{\textbf{Small}}} &
\multicolumn{3}{c}{\textcolor{blue}{\textbf{Tiny}}} \\
 & $\mathbf{AP}$ & $\mathbf{AP_{50}}$ & $\mathbf{AP_{25}}$ 
 & $\mathbf{AP}$ & $\mathbf{AP_{50}}$ & $\mathbf{AP_{25}}$ 
 & $\mathbf{AP}$ & $\mathbf{AP_{50}}$ & $\mathbf{AP_{25}}$ \\
\midrule
\multicolumn{10}{l}{\textbf{Closed-set}} \\
Mask3D \cite{mask3d} & \textbf{53.6} & \textbf{72.3} & \textbf{81.9} 
                     & 9.3           & 17.4          & 30.1             
                     & 4.2           & 13.5          & 16.8 \\
\midrule
\multicolumn{10}{l}{\textbf{Open-set}} \\
SAM3D~\cite{sam3d}   & 13.4          & 38.9          & 55.6          
                     & 8.5           & 17.2          & 28.3          
                     & 4.5           & 12.4          & 17.6 \\
\textbf{Ours}        & 28.5          & 48.1          & 68.5          
                     & \textbf{17.8} & \textbf{30.3} & \textbf{40.7} 
                     & \textbf{14.3} & \textbf{25.6} & \textbf{35.2} \\
\bottomrule
\end{tabular}
}
\vspace{-0.2cm}
\caption{Quantitative comparison with Mask3D (trained on ScanNet200) and SAM3D on our \textbf{ScanNet200-Fine50} test data across different instance sizes.}
\vspace{-0.2cm}
\label{tab:fine50_result}
\end{table}

\subsection{A Fine-grained Test Set -- \textit{ScanNet200-Fine50}}\label{sec:quant_finetest}
Nonetheless, we still cannot precisely evaluate our fine-grained predictions due to the lack of corresponding accurate GT annotations in available datasets.
To remedy this defect, we select 50 scenes from ScanNet200 \cite{scannet200} validation set, and provide high-quality and fine-grained annotations, yielding \textit{ScanNet200-Fine50} test set. A visual comparison between samples from the original ScanNet200 and our ScanNet200-Fine50 is provided in \cref{fig:fine50}. More samples are shown in the supplementary material.

Similar to \cite{segment3D}, we split ScanNet200-Fine50's annotations based on the mask size (\ie, number of points) of each instance to emphasize the fine-grained performance, including tiny ($0\sim1k$), small ($1\sim2k$), normal ($2k\sim\infty$).
As shown in \cref{tab:fine50_result}, although the closed-set method Mask3D \cite{mask3d} (trained on ScanNet200) surpasses ours on normal-sized instances, our method excels on tiny (12.1 $\mathbf{AP_{50}}$ $\uparrow$) and small objects (12.9$\uparrow$ $\mathbf{AP_{50}}$). In addition, our approach achieves superior results than the open-set method SAM3D \cite{sam3d} across all instance sizes.
These results are also supported by \cref{fig:qualitative_result}, validating the zero-shot ability of our method on \textit{highly fine-grained} segmentation. Our ScanNet200-Fine50 can serve as a supplementary test set to evaluate future zero-shot methods.

\begin{table}[t]
  \setlength{\tabcolsep}{3.0pt}
  \centering
  \resizebox{0.8\linewidth}{!}{
  \begin{tabular}{c|c|c|c|c}
    \toprule
    Method &SAM3D~\cite{sam3d} &Mask3D~\cite{mask3d} &Annotations~\cite{scannet200} & \textbf{Ours}\\
    \midrule
    mAcc  & 1.531$\pm$0.321 & 2.417$\pm$0.417 & 3.021$\pm$0.372 & \textbf{3.035$\pm$0.412}\\
    mDiv  & 1.552$\pm$0.491 & 2.308$\pm$0.420 & 2.909$\pm$0.409 & \textbf{2.927$\pm$0.361}\\
    \bottomrule
  \end{tabular}}
  \vspace{-0.2cm}
  \caption{The quantitative results of \textbf{user study}. “mAcc” and “mDiv” denote mean scores of accuracy and diversity.}
  \label{tab:user_study}
  \vspace{-0.4cm}
\end{table}

\subsection{User Study}\label{sec:user_study}
We further conduct a subjective user study following SAM \cite{kirillov2023sam}, as a complementary assessment.
We select a set of 20 reference results randomly, with most of them adjusted to a focused view for clearer discrimination.
We invited 100 subjects through an online questionnaire. All participants had no prior experience with 3D scene understanding, and none of them had seen our results before.
We present each subject with five images for each case, including qualitative results (on ScanNet200) of SAM3D, Mask3D, and ScanNet200's annotations, arranged side by side in random order, with an input image as a reference.
During the evaluation, we instruct subjects to rank the four results based on two criteria: segmentation \textit{accuracy} and \textit{diversity}. The accuracy evaluates the clarity of the segmented boundaries, while diversity focuses on the extent of whether “segmented anything”.
The result ranked first by the subjects is assigned a score of 4, while the last-ranked result receives 1.

The mean scores with standard deviation for accuracy (mAcc) and diversity (mDiv) are presented in \cref{tab:user_study}. The results are statistically signiﬁcant, demonstrating that our method surpasses both SAM3D and Mask3D by a large margin.
\textit{Notably}, even when compared to ScanNet200's annotations, our method achieves slightly higher scores in both quality and diversity. This user study further confirms the efficacy and zero-shot ability of our method.

\subsection{Ablation Studies and Analysis}\label{sec:ablation}

\paragraph{Efficiency.}

Our pipeline exhibits good efficiency, with the majority of computational time and memory usage allocated to the inference process of SAM across the RGB frames. In terms of memory usage, a single GPU with approximately 8000MB is sufficient to run SAM \cite{kirillov2023sam} along with our entire pipeline.

Regarding computational time, \cref{tab:time_cost} provides a breakdown of the time consumed by each step in our framework and compares ours with SAM3D \cite{sam3d}. Similar to SAM3D, our pipeline sequentially processes all frames, allowing room for speed improvement through parallel computation across frames.
In SAM3D, each 2D mask must be projected into 3D masks, which are then iteratively merged based on k-nearest-neighbor search across adjacent frames until achieving the final 3D segmentation of the entire scene. This iterative process increases the time cost.
As for Mask3D \cite{mask3d}, it needs $\sim$0.5s to perform inference on each scene, but requires $\sim$78 hours to train. Instead, our method does \textit{not} require training.

\begin{table}[t]
\setlength{\tabcolsep}{3.0pt}
  \centering
  \resizebox{0.56\linewidth}{!}{
  \begin{tabular}{cccc|c|c}
    \toprule
    Pro. & Sel. & Con. & Seg. & \textbf{Ours} (Total) & SAM3D \cite{sam3d}\\
    \midrule
    10 & 2 & 1 & 2 & \textbf{=15} & 20 \\
    \bottomrule
  \end{tabular}}
  \vspace{-0.2cm}
  \caption{\textbf{Running time} (in minutes) on the scene of $\sim$2,000 frames. “Pro.”, “Sel.”, “Con.” and “Seg.” respectively indicate prompt proposal, selection, consolidation, and 3D segmentation in our pipeline.}
  \label{tab:time_cost}
\end{table}

\begin{figure}[t]
\centering
    \centering
    \captionsetup{type=figure}
    \includegraphics[width=0.66\linewidth]{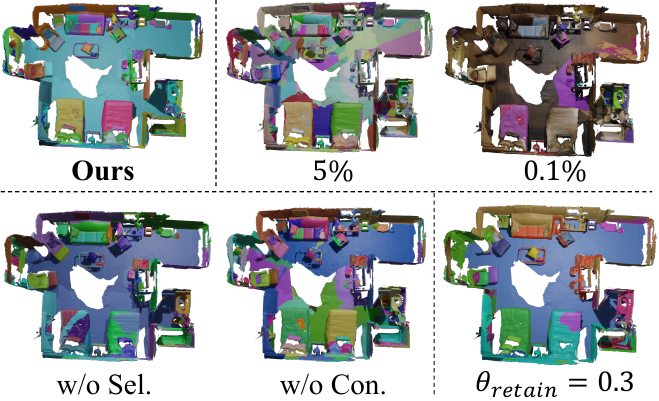}
    \vspace{-0.2cm}
    \captionof{figure}{The qualitative ablation studies on ScanNet200.} 
    \label{fig:ablation}
    \vspace{-0.2cm}
\end{figure}

\paragraph{Module impacts.}
As outlined in \cref{tab:quant_ablation} and depicted in \cref{fig:ablation}, removing View-Guided Prompt Selection results in a performance drop, and introduces inaccuracies in the examination of intersection areas during the prompt consolidation, thereby bringing fragmented segmentation.
On the other hand, omitting Surface-Based Prompt Consolidation conspicuously causes fragmented segmentation.

\begin{table}[t]
\centering
\setlength{\tabcolsep}{2.0pt}
\resizebox{1.0\linewidth}{!}{
\begin{tabular}{c|cc|cccc|cccccc|cc}
\toprule
& \multicolumn{2}{c|}{\textbf{Module Impact}} 
& \multicolumn{4}{c|}{\textbf{Initial Prompts}} 
& \multicolumn{6}{c|}{\textbf{$\mathtt{\theta}_{retain}$}}
& \multicolumn{2}{c}{\textbf{Selection}} \\
 & \textcolor{orange}{w/o Sel.} & \textcolor{orange}{w/o Con.}& 1\% & 1.5\% & 2\% & \textcolor{orange}{5\%} & \textcolor{orange}{0.3} & 0.4 & 0.5 & 0.6 & 0.7 & \textcolor{orange}{0.8} & soft & top-k \\
\midrule
$\mathbf{AP}$ & 19.5 & 21.4 & \textbf{26.3} & 25.8 & \textbf{26.3}
& 17.5 & 23.0 & \textbf{26.3} & 25.8 & \textbf{26.3} & 25.6 & 22.4 & 26.2 & 25.4 \\
$\mathbf{AP_{50}}$ & 40.8 & 41.2 & \textbf{47.2} & \textbf{47.2} & 46.9 & 39.8 & 43.0 & \textbf{47.2} & \textbf{47.2} & 46.9 & 46.2 & 42.3 & 47.0 & 47.1 \\
$\mathbf{AP_{25}}$ & 60.7 & 61.2 & \textbf{68.6} & 68.5 & \textbf{68.6} & 60.3 & 64.2 & \textbf{68.6} & \textbf{68.6} & 68.4 & 68.0 & 63.3 &
68.2 & 68.3\\
\bottomrule
\end{tabular}
}
\vspace{-0.2cm}
\caption{The quantitative ablation studies on ScanNet200. “w/o Sel.” and “w/o Con.” respectively denote discarding prompt selection and consolidation. We also evaluate our method using different ratios (1\%, 1.5\%, 2\%, 5\%) of input points as our initial prompts. $\mathtt{\theta}_{retain}$ is the threshold in prompt selection. “soft” and “top-k” are two voting schemes used during prompt selection.}
\label{tab:quant_ablation}
\end{table}

\vspace{-0.2cm}
\paragraph{The number of initial prompts.}
Our method performs stably well when an appropriate number of prompts is initialized (\cref{tab:quant_ablation}: 1\%, 1.5\%, 2\% of input points).
However, using too many initial prompts introduces more redundancy, which hurts the overall performance (\cref{tab:quant_ablation}: 5\%), and also causes inaccuracies in the examination of intersection areas during the prompt consolidation, thereby bringing fragmented segmentation (\cref{fig:ablation}: 5\%).
Using only a few initial prompts results in the absence of segmentation for many instances (\cref{fig:ablation}: 0.1\%).

\vspace{-0.2cm}
\paragraph{$\mathtt{\theta}_{retain}$ during prompt selection.}
During the View-Guided Prompt Selection (\cref{alg:prompt_filter}), we define a threshold $\mathtt{\theta}_{retain}$ to decide the probability of retaining a prompt.
Our method demonstrates satisfactory performance across a flexible range of $\mathtt{\theta}_{retain}$ (\cref{tab:quant_ablation}: $\mathtt{\theta}_{retain}=0.4, 0.5, 0.6, 0.7$).
Furthermore, setting $\mathtt{\theta}_{retain}$ to either a small or large value (\cref{tab:quant_ablation}: $\mathtt{\theta}_{retain}=0.3, 0.8$, \cref{fig:ablation}: $\mathtt{\theta}_{retain}=0.3$) \textit{weakens} the effect of prompt selection and causes performance drop, which conversely verifies the benefit of our selection algorithm.

\vspace{-0.3cm}
\paragraph{The way of prompt selection.}
As for our View-Guided Prompt Selection (\cref{alg:prompt_filter}), we also explored alternative selection methods.
One approach involved using the actual scores obtained during the selection process on individual frames and averaging these scores across all frames. We compared the mean score with $\mathtt{\theta}_{retain}$, to determine whether a prompt should be retained. This method can be referred to as “soft” voting.
Additionally, we tried another selection scheme where prompts are kept based on their frequency of being retained across all frames, referred to as the top-k voting scheme.
As shown in \cref{tab:quant_ablation}, both the “soft” and top-k voting schemes yield competitive results. This indicates the stability and effectiveness of our method in utilizing different selection ways.

\vspace{-0.3cm}
\paragraph{To summarize,} the above ablations demonstrate that our method does \textit{not} require a complex hyperparameter setup, highlighting its simplicity and effectiveness. The ablation results on our ScanNet200-Fine50 test set and more ablation studies are provided in the supplementary material.

\section{Conclusion}
\label{sec:conclusion}
We have proposed SAMPro3D for zero-shot segmentation of 3D scenes by utilizing SAM on 2D frames. The key idea is to locate SAM prompts in 3D to align pixel prompts across frames. 
Based on this key idea, we introduced a prompt selection algorithm and a prompt consolidation strategy to produce high-quality and comprehensive 3D segmentation.
Our method does not need any additional training, preserving the zero-shot capability of SAM.
\vspace{-0.2cm}
\paragraph{Limitations and future work.}
Our approach can not directly perform the multi-granularity segmentation like the original SAM. Future work may integrate user interactive prompts to explore the scene from different perspectives and levels of detail.
Our method can also be applied to NeRF \cite{mildenhall2020nerf} or 3DGS \cite{kerbl3Dgaussians}, since the key of our method is to locate SAM prompts in sparse point clouds, which can be reconstructed from novel view images of NeRF or 3DGS. Following this, our prompt selection and consolidation remain unchanged, directly yielding segmented novel views, omitting the point cloud segmentation stage.

Note that we aim to \textit{seamlessly transfer} the power of SAM-like models \textit{into 3D}. Addressing the limitation of SAM \textit{falls outside} our research. Future enhanced SAM-like models could be integrated into our model to improve 3D performance (we have tested HQ-SAM \cite{sam_hq} and Semantic-SAM \cite{li2023semanticsam} in the supplementary material).
Our method may not achieve perfect segmentation for all instances, primarily due to limitations in SAM's performance.


\section*{Acknowledgments}%
The work was supported in part by Guangdong Provincial Outstanding Youth Fund (No. 2023B1515020055), the Basic Research
Project No. HZQB-KCZYZ-2021067 of Hetao Shenzhen-HK S\&T
Cooperation Zone, the National Key R\&D Program of China with
grant No. 2018YFB1800800, Shenzhen Outstanding Talents Training
Fund 202002, Guangdong Research Projects No. 2017ZT07X152 and
No. 2019CX01X104, Key Area R\&D Program of Guangdong Province
(Grant No. 2018B030338001), the Guangdong Provincial Key Laboratory of Future Networks of Intelligence (Grant No. 2022B1212010001),
and Shenzhen Key Laboratory of Big Data and Artificial Intelligence
(Grant No. ZDSYS201707251409055). It is also partly supported by
NSFC-61931024, and Shenzhen Science and Technology Program
No. JCYJ20220530143604010.
{
    \small

    \bibliographystyle{ieeenat_fullname}

\begin{thebibliography}{96}
\providecommand{\natexlab}[1]{#1}
\providecommand{\url}[1]{\texttt{#1}}
\expandafter\ifx\csname urlstyle\endcsname\relax
  \providecommand{\doi}[1]{doi: #1}\else
  \providecommand{\doi}{doi: \begingroup \urlstyle{rm}\Url}\fi

\bibitem[{Armeni} et~al.(2016){Armeni}, {Sener}, {Zamir}, {Jiang}, {Brilakis}, {Fischer}, and {Savarese}]{s3dis}
I. {Armeni}, O. {Sener}, A.~R. {Zamir}, H. {Jiang}, I. {Brilakis}, M. {Fischer}, and S. {Savarese}.
\newblock 3d semantic parsing of large-scale indoor spaces.
\newblock In \emph{CVPR}, 2016.

\bibitem[Caesar et~al.(2020)Caesar, Bankiti, Lang, Vora, Liong, Xu, Krishnan, Pan, Baldan, and Beijbom]{nuscenes}
Holger Caesar, Varun Bankiti, Alex~H Lang, Sourabh Vora, Venice~Erin Liong, Qiang Xu, Anush Krishnan, Yu Pan, Giancarlo Baldan, and Oscar Beijbom.
\newblock nuscenes: A multimodal dataset for autonomous driving.
\newblock In \emph{CVPR}, 2020.

\bibitem[Caron et~al.(2021)Caron, Touvron, Misra, Jégou, Mairal, Bojanowski, and Joulin]{caron2021dino}
Mathilde Caron, Hugo Touvron, Ishan Misra, Hervé Jégou, Julien Mairal, Piotr Bojanowski, and Armand Joulin.
\newblock Emerging properties in self-supervised vision transformers.
\newblock In \emph{ICCV}, 2021.

\bibitem[Cen et~al.(2023)Cen, Zhou, Fang, Yang, Shen, Xie, Jiang, Zhang, and Tian]{sa3d}
Jiazhong Cen, Zanwei Zhou, Jiemin Fang, Chen Yang, Wei Shen, Lingxi Xie, Dongsheng Jiang, Xiaopeng Zhang, and Qi Tian.
\newblock Segment anything in 3d with nerfs.
\newblock In \emph{NeurIPS}, 2023.

\bibitem[Chang et~al.(2017)Chang, Dai, Funkhouser, Halber, Niessner, Savva, Song, Zeng, and Zhang]{matterport3D}
Angel Chang, Angela Dai, Thomas Funkhouser, Maciej Halber, Matthias Niessner, Manolis Savva, Shuran Song, Andy Zeng, and Yinda Zhang.
\newblock Matterport3d: Learning from rgb-d data in indoor environments.
\newblock In \emph{3DV}, 2017.

\bibitem[Chen et~al.(2020)Chen, Chang, and Nie{\ss}ner]{chen2020scanrefer}
Dave~Zhenyu Chen, Angel~X Chang, and Matthias Nie{\ss}ner.
\newblock Scanrefer: 3d object localization in rgb-d scans using natural language.
\newblock In \emph{ECCV}, 2020.

\bibitem[Chen et~al.(2023{\natexlab{a}})Chen, Liu, Kong, Zhu, Ma, Li, Hou, Qiao, and Wang]{chen2023clip2scene}
Runnan Chen, Youquan Liu, Lingdong Kong, Xinge Zhu, Yuexin Ma, Yikang Li, Yuenan Hou, Yu Qiao, and Wenping Wang.
\newblock Clip2scene: Towards label-efficient 3d scene understanding by clip.
\newblock In \emph{CVPR}, 2023{\natexlab{a}}.

\bibitem[Chen et~al.(2022)Chen, Nie{\ss}ner, and Dai]{4dcontrast}
Yujin Chen, Matthias Nie{\ss}ner, and Angela Dai.
\newblock 4dcontrast: Contrastive learning with dynamic correspondences for 3d scene understanding.
\newblock In \emph{ECCV}, 2022.

\bibitem[Chen et~al.(2023{\natexlab{b}})Chen, Liu, Zhang, Qi, and Jia]{chen2023voxenext}
Yukang Chen, Jianhui Liu, Xiangyu Zhang, Xiaojuan Qi, and Jiaya Jia.
\newblock Voxelnext: Fully sparse voxelnet for 3d object detection and tracking.
\newblock In \emph{CVPR}, 2023{\natexlab{b}}.

\bibitem[Cheng et~al.(2021)Cheng, Hui, Xie, and Yang]{cheng2021sspc}
Mingmei Cheng, Le Hui, Jin Xie, and Jian Yang.
\newblock Sspc-net: Semi-supervised semantic 3d point cloud segmentation network.
\newblock In \emph{AAAI}, 2021.

\bibitem[Chibane et~al.(2022)Chibane, Engelmann, Tran, and Pons-Moll]{chibane2021box2mask}
Julian Chibane, Francis Engelmann, Tuan~Anh Tran, and Gerard Pons-Moll.
\newblock Box2mask: Weakly supervised 3d semantic instance segmentation using bounding boxes.
\newblock In \emph{ECCV}, 2022.

\bibitem[Choy et~al.(2019)Choy, Gwak, and Savarese]{minkowski}
Christopher Choy, JunYoung Gwak, and Silvio Savarese.
\newblock {4D Spatio-Temporal ConvNets: Minkowski Convolutional Neural Networks}.
\newblock In \emph{CVPR}, 2019.

\bibitem[Dai et~al.(2017{\natexlab{a}})Dai, Chang, Savva, Halber, Funkhouser, and Nie{\ss}ner]{scannet}
Angela Dai, Angel~X. Chang, Manolis Savva, Maciej Halber, Thomas Funkhouser, and Matthias Nie{\ss}ner.
\newblock Scannet: Richly-annotated 3d reconstructions of indoor scenes.
\newblock In \emph{CVPR}, 2017{\natexlab{a}}.

\bibitem[Dai et~al.(2017{\natexlab{b}})Dai, Nie{\ss}ner, Zoll{\"o}fer, Izadi, and Theobalt]{dai2017bundlefusion}
Angela Dai, Matthias Nie{\ss}ner, Michael Zoll{\"o}fer, Shahram Izadi, and Christian Theobalt.
\newblock Bundlefusion: Real-time globally consistent 3d reconstruction using on-the-fly surface re-integration.
\newblock \emph{TOG}, 2017{\natexlab{b}}.

\bibitem[Ding et~al.(2023)Ding, Yang, Xue, Zhang, Bai, and Qi]{pla}
Runyu Ding, Jihan Yang, Chuhui Xue, Wenqing Zhang, Song Bai, and Xiaojuan Qi.
\newblock {PLA: Language-Driven Open-Vocabulary 3D Scene Understanding}.
\newblock In \emph{CVPR}, 2023.

\bibitem[Ding et~al.(2024)Ding, Yang, Xue, Zhang, Bai, and Qi]{lowis3d}
Runyu Ding, Jihan Yang, Chuhui Xue, Wenqing Zhang, Song Bai, and Xiaojuan Qi.
\newblock Lowis3d: Language-driven open-world instance-level 3d scene understanding.
\newblock \emph{PAMI}, 2024.

\bibitem[Dong et~al.(2023)Dong, Liu, and Lin]{dong2023leveraginglargescalepretrainedvision}
Shichao Dong, Fayao Liu, and Guosheng Lin.
\newblock Leveraging large-scale pretrained vision foundation models for label-efficient 3d point cloud segmentation.
\newblock \emph{arXiv preprint arXiv:2311.01989}, 2023.

\bibitem[Felzenszwalb and Huttenlocher(2004)]{felzenszwalb2004efficient}
Pedro~F Felzenszwalb and Daniel~P Huttenlocher.
\newblock Efficient graph-based image segmentation.
\newblock \emph{IJCV}, 2004.

\bibitem[Geiger et~al.(2012)Geiger, Lenz, and Urtasun]{kitti}
Andreas Geiger, Philip Lenz, and Raquel Urtasun.
\newblock Are we ready for autonomous driving? the kitti vision benchmark suite.
\newblock In \emph{CVPR}, 2012.

\bibitem[Genova et~al.(2021)Genova, Yin, Kundu, Pantofaru, Cole, Sud, Brewington, Shucker, and Funkhouser]{genova2021learning}
Kyle Genova, Xiaoqi Yin, Abhijit Kundu, Caroline Pantofaru, Forrester Cole, Avneesh Sud, Brian Brewington, Brian Shucker, and Thomas Funkhouser.
\newblock Learning 3d semantic segmentation with only 2d image supervision.
\newblock In \emph{3DV}, 2021.

\bibitem[Graham et~al.(2018)Graham, Engelcke, and van~der Maaten]{sparseconv}
Benjamin Graham, Martin Engelcke, and Laurens van~der Maaten.
\newblock 3d semantic segmentation with submanifold sparse convolutional networks.
\newblock In \emph{CVPR}, 2018.

\bibitem[Guo et~al.(2024)Guo, Zhu, Peng, Wang, Shen, Hu, and Zhou]{samgraphcut}
Haoyu Guo, He Zhu, Sida Peng, Yuang Wang, Yujun Shen, Ruizhen Hu, and Xiaowei Zhou.
\newblock Sam-guided graph cut for 3d instance segmentation.
\newblock In \emph{ECCV}, 2024.

\bibitem[Gupta et~al.(2019)Gupta, Dollar, and Girshick]{gupta2019lvis}
Agrim Gupta, Piotr Dollar, and Ross Girshick.
\newblock Lvis: A dataset for large vocabulary instance segmentation.
\newblock In \emph{CVPR}, 2019.

\bibitem[Hou et~al.(2021)Hou, Graham, Nie{\ss}ner, and Xie]{csc}
Ji Hou, Benjamin Graham, Matthias Nie{\ss}ner, and Saining Xie.
\newblock Exploring data-efficient 3d scene understanding with contrastive scene contexts.
\newblock In \emph{CVPR}, 2021.

\bibitem[Hua et~al.(2018)Hua, Tran, and Yeung]{pointwisecnn}
Binh{-}Son Hua, Minh{-}Khoi Tran, and Sai{-}Kit Yeung.
\newblock {Point-wise Convolutional Neural Network}.
\newblock In \emph{CVPR}, 2018.

\bibitem[Hua et~al.(2016)Hua, Pham, Nguyen, Tran, Yu, and Yeung]{scenenn}
Binh-Son Hua, Quang-Hieu Pham, Duc~Thanh Nguyen, Minh-Khoi Tran, Lap-Fai Yu, and Sai-Kit Yeung.
\newblock Scenenn: A scene meshes dataset with annotations.
\newblock In \emph{3DV}, 2016.

\bibitem[Huang et~al.(2023)Huang, Peng, He, Yang, Zhou, and Ouyang]{huang2023ponder}
Di Huang, Sida Peng, Tong He, Honghui Yang, Xiaowei Zhou, and Wanli Ouyang.
\newblock Ponder: Point cloud pre-training via neural rendering.
\newblock In \emph{ICCV}, 2023.

\bibitem[Huang et~al.(2024{\natexlab{a}})Huang, Peng, Takmaz, Tombari, Pollefeys, Song, Huang, and Engelmann]{segment3D}
Rui Huang, Songyou Peng, Ayca Takmaz, Federico Tombari, Marc Pollefeys, Shiji Song, Gao Huang, and Francis Engelmann.
\newblock Segment3d: Learning fine-grained class-agnostic 3d segmentation without manual labels.
\newblock In \emph{ECCV}, 2024{\natexlab{a}}.

\bibitem[Huang et~al.(2021)Huang, Xie, Zhu, and Zhu]{huang2021spatio}
Siyuan Huang, Yichen Xie, Song-Chun Zhu, and Yixin Zhu.
\newblock Spatio-temporal self-supervised representation learning for 3d point clouds.
\newblock In \emph{ICCV}, 2021.

\bibitem[Huang et~al.(2024{\natexlab{b}})Huang, Wu, Chen, Zhao, Zhu, and Lasenby]{openins3d}
Zhening Huang, Xiaoyang Wu, Xi Chen, Hengshuang Zhao, Lei Zhu, and Joan Lasenby.
\newblock Openins3d: Snap and lookup for 3d open-vocabulary instance segmentation.
\newblock In \emph{ECCV}, 2024{\natexlab{b}}.

\bibitem[Jiang et~al.(2019)Jiang, Zhao, Liu, Shen, Fu, and Jia]{pointweb}
Li Jiang, Hengshuang Zhao, Shu Liu, Xiaoyong Shen, Chi-Wing Fu, and Jiaya Jia.
\newblock Hierarchical point-edge interaction network for point cloud semantic segmentation.
\newblock In \emph{ICCV}, 2019.

\bibitem[Jiang et~al.(2020)Jiang, Zhao, Shi, Liu, Fu, and Jia]{pointgroup}
Li Jiang, Hengshuang Zhao, Shaoshuai Shi, Shu Liu, Chi-Wing Fu, and Jiaya Jia.
\newblock Pointgroup: Dual-set point grouping for 3d instance segmentation.
\newblock In \emph{CVPR}, 2020.

\bibitem[Ke et~al.(2023)Ke, Ye, Danelljan, Liu, Tai, Tang, and Yu]{sam_hq}
Lei Ke, Mingqiao Ye, Martin Danelljan, Yifan Liu, Yu-Wing Tai, Chi-Keung Tang, and Fisher Yu.
\newblock Segment anything in high quality.
\newblock In \emph{NeurIPS}, 2023.

\bibitem[Kerbl et~al.(2023)Kerbl, Kopanas, Leimk{\"u}hler, and Drettakis]{kerbl3Dgaussians}
Bernhard Kerbl, Georgios Kopanas, Thomas Leimk{\"u}hler, and George Drettakis.
\newblock 3d gaussian splatting for real-time radiance field rendering.
\newblock \emph{TOG}, 2023.

\bibitem[Kirillov et~al.(2023)Kirillov, Mintun, Ravi, Mao, Rolland, Gustafson, Xiao, Whitehead, Berg, Lo, Dollár, and Girshick]{kirillov2023sam}
Alexander Kirillov, Eric Mintun, Nikhila Ravi, Hanzi Mao, Chloe Rolland, Laura Gustafson, Tete Xiao, Spencer Whitehead, Alexander~C. Berg, Wan-Yen Lo, Piotr Dollár, and Ross Girshick.
\newblock Segment anything.
\newblock In \emph{ICCV}, 2023.

\bibitem[Kundu et~al.(2020)Kundu, Yin, Fathi, Ross, Brewington, Funkhouser, and Pantofaru]{kundu2020virtual}
Abhijit Kundu, Xiaoqi Yin, Alireza Fathi, David Ross, Brian Brewington, Thomas Funkhouser, and Caroline Pantofaru.
\newblock Virtual multi-view fusion for 3d semantic segmentation.
\newblock In \emph{ECCV}, 2020.

\bibitem[Lambert et~al.(2020)Lambert, Liu, Sener, Hays, and Koltun]{MSeg_2020_CVPR}
John Lambert, Zhuang Liu, Ozan Sener, James Hays, and Vladlen Koltun.
\newblock {MSeg}: A composite dataset for multi-domain semantic segmentation.
\newblock In \emph{CVPR}, 2020.

\bibitem[Li et~al.(2023)Li, Zhang, Sun, Zou, Liu, Yang, Li, Zhang, and Gao]{li2023semanticsam}
Feng Li, Hao Zhang, Peize Sun, Xueyan Zou, Shilong Liu, Jianwei Yang, Chunyuan Li, Lei Zhang, and Jianfeng Gao.
\newblock Semantic-sam: Segment and recognize anything at any granularity.
\newblock \emph{arXiv preprint arXiv:2307.04767}, 2023.

\bibitem[Li et~al.(2018)Li, Bu, Sun, Wu, Di, and Chen]{pointcnn}
Yangyan Li, Rui Bu, Mingchao Sun, Wei Wu, Xinhan Di, and Baoquan Chen.
\newblock {PointCNN: Convolution on X-transformed Points}.
\newblock In \emph{NeurIPS}, 2018.

\bibitem[Li et~al.(2022)Li, Mao, Girshick, and He]{vitdet}
Yanghao Li, Hanzi Mao, Ross Girshick, and Kaiming He.
\newblock Exploring plain vision transformer backbones for object detection.
\newblock In \emph{ECCV}, 2022.

\bibitem[Li et~al.(2021)Li, Yu, Sang, Wang, Song, Liu, Yeh, Zhu, Gundavarapu, Shi, Bi, Yu, Xu, Sunkavalli, Hasan, Ramamoorthi, and Chandraker]{openroom}
Zhengqin Li, Ting-Wei Yu, Shen Sang, Sarah Wang, Meng Song, Yuhan Liu, Yu-Ying Yeh, Rui Zhu, Nitesh Gundavarapu, Jia Shi, Sai Bi, Hong-Xing Yu, Zexiang Xu, Kalyan Sunkavalli, Milos Hasan, Ravi Ramamoorthi, and Manmohan Chandraker.
\newblock Openrooms: An open framework for photorealistic indoor scene datasets.
\newblock In \emph{CVPR}, 2021.

\bibitem[Liao et~al.(2022)Liao, Xie, and Geiger]{kitti360}
Yiyi Liao, Jun Xie, and Andreas Geiger.
\newblock {KITTI}-360: A novel dataset and benchmarks for urban scene understanding in 2d and 3d.
\newblock \emph{PAMI}, 2022.

\bibitem[Lin et~al.(2014)Lin, Maire, Belongie, Hays, Perona, Ramanan, Doll{\'a}r, and Zitnick]{coco}
Tsung-Yi Lin, Michael Maire, Serge Belongie, James Hays, Pietro Perona, Deva Ramanan, Piotr Doll{\'a}r, and C~Lawrence Zitnick.
\newblock Microsoft coco: Common objects in context.
\newblock In \emph{ECCV}, 2014.

\bibitem[Lin et~al.(2020)Lin, Yan, Huang, Du, Liu, Cui, and Han]{fpconv}
Yiqun Lin, Zizheng Yan, Haibin Huang, Dong Du, Ligang Liu, Shuguang Cui, and Xiaoguang Han.
\newblock Fpconv: Learning local flattening for point convolution.
\newblock In \emph{CVPR}, 2020.

\bibitem[Liu et~al.(2022)Liu, Zhao, Nie, Gao, and Chen]{liu2022weakly}
Kangcheng Liu, Yuzhi Zhao, Qiang Nie, Zhi Gao, and Ben~M Chen.
\newblock Weakly supervised 3d scene segmentation with region-level boundary awareness and instance discrimination.
\newblock In \emph{ECCV}, 2022.

\bibitem[Liu et~al.(2023)Liu, Kong, Cen, Chen, Zhang, Pan, Chen, and Liu]{liu2023segment}
Youquan Liu, Lingdong Kong, Jun Cen, Runnan Chen, Wenwei Zhang, Liang Pan, Kai Chen, and Ziwei Liu.
\newblock Segment any point cloud sequences by distilling vision foundation models.
\newblock In \emph{NeurIPS}, 2023.

\bibitem[Liu et~al.(2020)Liu, Hu, Cao, Zhang, and Tong]{closerlook}
Ze Liu, Han Hu, Yue Cao, Zheng Zhang, and Xin Tong.
\newblock A closer look at local aggregation operators in point cloud analysis.
\newblock In \emph{ECCV}, 2020.

\bibitem[Liu et~al.(2021)Liu, Zhang, Cao, Hu, and Tong]{groupfree}
Ze Liu, Zheng Zhang, Yue Cao, Han Hu, and Xin Tong.
\newblock Group-free 3d object detection via transformers.
\newblock In \emph{ICCV}, 2021.

\bibitem[Lu et~al.(2023)Lu, Chang, Jing, Boularias, and Bekris]{ovir3d}
Shiyang Lu, Haonan Chang, Eric~Pu Jing, Abdeslam Boularias, and Kostas Bekris.
\newblock Ovir-3d: Open-vocabulary 3d instance retrieval without training on 3d data.
\newblock In \emph{CoRL}, 2023.

\bibitem[Ma et~al.(2022)Ma, Qin, You, Ran, and Fu]{ma2022rethinking}
Xu Ma, Can Qin, Haoxuan You, Haoxi Ran, and Yun Fu.
\newblock Rethinking network design and local geometry in point cloud: A simple residual mlp framework.
\newblock In \emph{ICLR}, 2022.

\bibitem[McCormac et~al.(2017)McCormac, Handa, Davison, and Leutenegger]{mccormac2017semanticfusion}
John McCormac, Ankur Handa, Andrew Davison, and Stefan Leutenegger.
\newblock Semanticfusion: Dense 3d semantic mapping with convolutional neural networks.
\newblock In \emph{ICRA}, 2017.

\bibitem[McInnes and Healy(2017)]{hdbscan}
Leland McInnes and John Healy.
\newblock Accelerated hierarchical density based clustering.
\newblock In \emph{ICDMW}, 2017.

\bibitem[Michele et~al.(2021)Michele, Boulch, Puy, Bucher, and Marlet]{michele2021generative}
Bj{\"o}rn Michele, Alexandre Boulch, Gilles Puy, Maxime Bucher, and Renaud Marlet.
\newblock Generative zero-shot learning for semantic segmentation of {3D} point cloud.
\newblock In \emph{3DV}, 2021.

\bibitem[Mildenhall et~al.(2020)Mildenhall, Srinivasan, Tancik, Barron, Ramamoorthi, and Ng]{mildenhall2020nerf}
Ben Mildenhall, Pratul~P. Srinivasan, Matthew Tancik, Jonathan~T. Barron, Ravi Ramamoorthi, and Ren Ng.
\newblock Nerf: Representing scenes as neural radiance fields for view synthesis.
\newblock In \emph{ECCV}, 2020.

\bibitem[Misra et~al.(2021)Misra, Girdhar, and Joulin]{3detr}
Ishan Misra, Rohit Girdhar, and Armand Joulin.
\newblock {An End-to-End Transformer Model for 3D Object Detection}.
\newblock In \emph{ICCV}, 2021.

\bibitem[Nguyen et~al.(2024)Nguyen, Ngo, Gan, Kalogerakis, Tran, Pham, and Nguyen]{open3dis}
Phuc D.~A. Nguyen, Tuan~Duc Ngo, Chuang Gan, Evangelos Kalogerakis, Anh Tran, Cuong Pham, and Khoi Nguyen.
\newblock Open3dis: Open-vocabulary 3d instance segmentation with 2d mask guidance.
\newblock In \emph{CVPR}, 2024.

\bibitem[Oquab et~al.(2023)Oquab, Darcet, Moutakanni, Vo, Szafraniec, Khalidov, Fernandez, Haziza, Massa, El-Nouby, Assran, Ballas, Galuba, Howes, Huang, Li, Misra, Rabbat, Sharma, Synnaeve, Xu, Jegou, Mairal, Labatut, Joulin, and Bojanowski]{oquab2023dinov2}
Maxime Oquab, Timothée Darcet, Théo Moutakanni, Huy Vo, Marc Szafraniec, Vasil Khalidov, Pierre Fernandez, Daniel Haziza, Francisco Massa, Alaaeldin El-Nouby, Mahmoud Assran, Nicolas Ballas, Wojciech Galuba, Russell Howes, Po-Yao Huang, Shang-Wen Li, Ishan Misra, Michael Rabbat, Vasu Sharma, Gabriel Synnaeve, Hu Xu, Hervé Jegou, Julien Mairal, Patrick Labatut, Armand Joulin, and Piotr Bojanowski.
\newblock Dinov2: Learning robust visual features without supervision.
\newblock \emph{arXiv preprint arXiv:2304.07193}, 2023.

\bibitem[Paszke et~al.(2019)Paszke, Gross, Massa, Lerer, Bradbury, Chanan, Killeen, Lin, Gimelshein, Antiga, Desmaison, Kopf, Yang, DeVito, Raison, Tejani, Chilamkurthy, Steiner, Fang, Bai, and Chintala]{pytorch}
Adam Paszke, Sam Gross, Francisco Massa, Adam Lerer, James Bradbury, Gregory Chanan, Trevor Killeen, Zeming Lin, Natalia Gimelshein, Luca Antiga, Alban Desmaison, Andreas Kopf, Edward Yang, Zachary DeVito, Martin Raison, Alykhan Tejani, Sasank Chilamkurthy, Benoit Steiner, Lu Fang, Junjie Bai, and Soumith Chintala.
\newblock Pytorch: An imperative style, high-performance deep learning library.
\newblock In \emph{NeurIPS}, 2019.

\bibitem[Peng et~al.(2023)Peng, Genova, Jiang, Tagliasacchi, Pollefeys, and Funkhouser]{openscene}
Songyou Peng, Kyle Genova, Chiyu~"Max" Jiang, Andrea Tagliasacchi, Marc Pollefeys, and Thomas Funkhouser.
\newblock {OpenScene: 3D Scene Understanding with Open Vocabularies}.
\newblock In \emph{CVPR}, 2023.

\bibitem[Qi et~al.(2017{\natexlab{a}})Qi, Su, Mo, and Guibas]{pointnet}
Charles~R. Qi, Hao Su, Kaichun Mo, and Leonidas~J. Guibas.
\newblock {PointNet: Deep Learning on Point Sets for 3D Classification and Segmentation}.
\newblock In \emph{CVPR}, 2017{\natexlab{a}}.

\bibitem[Qi et~al.(2017{\natexlab{b}})Qi, Yi, Su, and Guibas]{pointnet++}
Charles~R. Qi, Li Yi, Hao Su, and Leonidas~J. Guibas.
\newblock {PointNet++: Deep Hierarchical Feature Learning on Point Sets in a Metric Space}.
\newblock In \emph{NeurIPS}, 2017{\natexlab{b}}.

\bibitem[Qi et~al.(2019)Qi, Litany, He, and Guibas]{votenet}
Charles~R Qi, Or Litany, Kaiming He, and Leonidas~J Guibas.
\newblock Deep hough voting for 3d object detection in point clouds.
\newblock In \emph{ICCV}, 2019.

\bibitem[Radford et~al.(2021)Radford, Kim, Hallacy, Ramesh, Goh, Agarwal, Sastry, Askell, Mishkin, Clark, Krueger, and Sutskever]{radford2021clip}
Alec Radford, Jong~Wook Kim, Chris Hallacy, Aditya Ramesh, Gabriel Goh, Sandhini Agarwal, Girish Sastry, Amanda Askell, Pamela Mishkin, Jack Clark, Gretchen Krueger, and Ilya Sutskever.
\newblock Learning transferable visual models from natural language supervision.
\newblock In \emph{ICML}, 2021.

\bibitem[Rajič et~al.(2023)Rajič, Ke, Tai, Tang, Danelljan, and Yu]{sam-pt}
Frano Rajič, Lei Ke, Yu-Wing Tai, Chi-Keung Tang, Martin Danelljan, and Fisher Yu.
\newblock Segment anything meets point tracking.
\newblock \emph{arXiv preprint arXiv:2307.01197}, 2023.

\bibitem[Ren et~al.(2015)Ren, He, Girshick, and Sun]{fasterrcnn}
Shaoqing Ren, Kaiming He, Ross Girshick, and Jian Sun.
\newblock Faster r-cnn: Towards real-time object detection with region proposal networks.
\newblock In \emph{NeurIPS}, 2015.

\bibitem[Rozenberszki et~al.(2022)Rozenberszki, Litany, and Dai]{scannet200}
David Rozenberszki, Or Litany, and Angela Dai.
\newblock Language-grounded indoor 3d semantic segmentation in the wild.
\newblock In \emph{ECCV}, 2022.

\bibitem[Rozenberszki et~al.(2024)Rozenberszki, Litany, and Dai]{unscene3d}
David Rozenberszki, Or Litany, and Angela Dai.
\newblock Unscene3d: Unsupervised 3d instance segmentation for indoor scenes.
\newblock In \emph{CVPR}, 2024.

\bibitem[Sautier et~al.(2022)Sautier, Puy, Gidaris, Boulch, Bursuc, and Marlet]{sautier2022image}
Corentin Sautier, Gilles Puy, Spyros Gidaris, Alexandre Boulch, Andrei Bursuc, and Renaud Marlet.
\newblock Image-to-lidar self-supervised distillation for autonomous driving data.
\newblock In \emph{CVPR}, 2022.

\bibitem[Schult et~al.(2023)Schult, Engelmann, Hermans, Litany, Tang, and Leibe]{mask3d}
Jonas Schult, Francis Engelmann, Alexander Hermans, Or Litany, Siyu Tang, and Bastian Leibe.
\newblock {Mask3D: Mask Transformer for 3D Semantic Instance Segmentation}.
\newblock In \emph{ICRA}, 2023.

\bibitem[Shi et~al.(2019)Shi, Wang, and Li]{PointRCNN}
Shaoshuai Shi, Xiaogang Wang, and Hongsheng Li.
\newblock Pointrcnn: 3d object proposal generation and detection from point cloud.
\newblock In \emph{CVPR}, 2019.

\bibitem[Song et~al.(2015)Song, Lichtenberg, and Xiao]{sunrgbd}
Shuran Song, Samuel~P. Lichtenberg, and Jianxiong Xiao.
\newblock Sun rgb-d: A rgb-d scene understanding benchmark suite.
\newblock In \emph{CVPR}, 2015.

\bibitem[Sun et~al.(2020)Sun, Kretzschmar, Dotiwalla, Chouard, Patnaik, Tsui, Guo, Zhou, Chai, Caine, et~al.]{waymo}
Pei Sun, Henrik Kretzschmar, Xerxes Dotiwalla, Aurelien Chouard, Vijaysai Patnaik, Paul Tsui, James Guo, Yin Zhou, Yuning Chai, Benjamin Caine, et~al.
\newblock Scalability in perception for autonomous driving: Waymo open dataset.
\newblock In \emph{CVPR}, 2020.

\bibitem[Takmaz et~al.(2023)Takmaz, Fedele, Sumner, Pollefeys, Tombari, and Engelmann]{openmask3d}
Ay{\c{c}}a Takmaz, Elisabetta Fedele, Robert~W. Sumner, Marc Pollefeys, Federico Tombari, and Francis Engelmann.
\newblock {OpenMask3D: Open-Vocabulary 3D Instance Segmentation}.
\newblock In \emph{NeurIPS}, 2023.

\bibitem[Thomas et~al.(2019)Thomas, Qi, Deschaud, Marcotegui, Goulette, and Guibas]{kpconv}
Hugues Thomas, Charles~R. Qi, Jean-Emmanuel Deschaud, Beatriz Marcotegui, Fran{\c{c}}ois Goulette, and Leonidas~J. Guibas.
\newblock Kpconv: Flexible and deformable convolution for point clouds.
\newblock In \emph{ICCV}, 2019.

\bibitem[Vineet et~al.(2015)Vineet, Miksik, Lidegaard, Nie{\ss}ner, Golodetz, Prisacariu, K{\"a}hler, Murray, Izadi, P{\'e}rez, et~al.]{vineet2015incremental}
Vibhav Vineet, Ondrej Miksik, Morten Lidegaard, Matthias Nie{\ss}ner, Stuart Golodetz, Victor~A Prisacariu, Olaf K{\"a}hler, David~W Murray, Shahram Izadi, Patrick P{\'e}rez, et~al.
\newblock Incremental dense semantic stereo fusion for large-scale semantic scene reconstruction.
\newblock In \emph{ICRA}, 2015.

\bibitem[Wang et~al.(2022{\natexlab{a}})Wang, Yang, Men, Lin, Bai, Li, Ma, Zhou, Zhou, and Yang]{wang2022ofa}
Peng Wang, An Yang, Rui Men, Junyang Lin, Shuai Bai, Zhikang Li, Jianxin Ma, Chang Zhou, Jingren Zhou, and Hongxia Yang.
\newblock Ofa: Unifying architectures, tasks, and modalities through a simple sequence-to-sequence learning framework.
\newblock \emph{CoRR}, 2022{\natexlab{a}}.

\bibitem[Wang et~al.(2023)Wang, Zhang, Cao, Wang, Shen, and Huang]{wang2023segGPT}
Xinlong Wang, Xiaosong Zhang, Yue Cao, Wen Wang, Chunhua Shen, and Tiejun Huang.
\newblock Seggpt: Segmenting everything in context.
\newblock In \emph{ICCV}, 2023.

\bibitem[Wang et~al.(2019)Wang, Sun, Liu, Sarma, Bronstein, and Solomon]{dgcnn}
Yue Wang, Yongbin Sun, Ziwei Liu, Sanjay~E. Sarma, Michael~M. Bronstein, and Justin~M. Solomon.
\newblock Dynamic graph cnn for learning on point clouds.
\newblock \emph{ACM Trans. Graph.}, 2019.

\bibitem[Wang et~al.(2022{\natexlab{b}})Wang, Guizilini, Zhang, Wang, Zhao, and Solomon]{wang2022detr3d}
Yue Wang, Vitor~Campagnolo Guizilini, Tianyuan Zhang, Yilun Wang, Hang Zhao, and Justin Solomon.
\newblock Detr3d: 3d object detection from multi-view images via 3d-to-2d queries.
\newblock In \emph{CoRL}, 2022{\natexlab{b}}.

\bibitem[Wu et~al.(2019)Wu, Qi, and Fuxin]{wu2019pointconv}
Wenxuan Wu, Zhongang Qi, and Li Fuxin.
\newblock Pointconv: Deep convolutional networks on 3d point clouds.
\newblock In \emph{CVPR}, 2019.

\bibitem[Xie et~al.(2020)Xie, Gu, Guo, Qi, Guibas, and Litany]{PointContrast}
Saining Xie, Jiatao Gu, Demi Guo, Charles~R. Qi, Leonidas Guibas, and Or Litany.
\newblock Pointcontrast: Unsupervised pre-training for 3d point cloud understanding.
\newblock In \emph{ECCV}, 2020.

\bibitem[Xu et~al.(2021)Xu, Ding, Zhao, and Qi]{paconv}
Mutian Xu, Runyu Ding, Hengshuang Zhao, and Xiaojuan Qi.
\newblock Paconv: Position adaptive convolution with dynamic kernel assembling on point clouds.
\newblock In \emph{CVPR}, 2021.

\bibitem[Xu et~al.(2022)Xu, Chen, Liu, and Han]{toscene}
Mutian Xu, Pei Chen, Haolin Liu, and Xiaoguang Han.
\newblock To-scene: A large-scale dataset for understanding 3d tabletop scenes.
\newblock In \emph{ECCV}, 2022.

\bibitem[Xu et~al.(2023)Xu, Xu, He, Ouyang, Wang, Han, and Qiao]{xu2022mm3Dscene}
Mingye Xu, Mutian Xu, Tong He, Wanli Ouyang, Yali Wang, Xiaoguang Han, and Yu Qiao.
\newblock Mm-3dscene: 3d scene understanding by customizing masked modeling with informative-preserved reconstruction and self-distilled consistency.
\newblock In \emph{CVPR}, 2023.

\bibitem[Xu and Lee(2020)]{XuLee_CVPR20}
Xun Xu and Gim~Hee Lee.
\newblock Weakly supervised semantic point cloud segmentation: Towards 10x fewer labels.
\newblock In \emph{CVPR}, 2020.

\bibitem[Xu et~al.(2018)Xu, Fan, Xu, Zeng, and Qiao]{spidercnn}
Yifan Xu, Tianqi Fan, Mingye Xu, Long Zeng, and Yu Qiao.
\newblock {SpiderCNN: Deep Learning on Point Sets with Parameterized Convolutional Filters}.
\newblock In \emph{ECCV}, 2018.

\bibitem[Yang et~al.(2023{\natexlab{a}})Yang, Hayat, Jin, Zhu, and Lei]{yang2023zero}
Yuwei Yang, Munawar Hayat, Zhao Jin, Hongyuan Zhu, and Yinjie Lei.
\newblock Zero-shot point cloud segmentation by semantic-visual aware synthesis.
\newblock In \emph{ICCV}, 2023{\natexlab{a}}.

\bibitem[Yang et~al.(2023{\natexlab{b}})Yang, Wu, He, Zhao, and Liu]{sam3d}
Yunhan Yang, Xiaoyang Wu, Tong He, Hengshuang Zhao, and Xihui Liu.
\newblock Sam3d: Segment anything in 3d scenes.
\newblock In \emph{ICCVW}, 2023{\natexlab{b}}.

\bibitem[Yeshwanth et~al.(2023)Yeshwanth, Liu, Nie{\ss}ner, and Dai]{scannetpp}
Chandan Yeshwanth, Yueh-Cheng Liu, Matthias Nie{\ss}ner, and Angela Dai.
\newblock Scannet++: A high-fidelity dataset of 3d indoor scenes.
\newblock In \emph{ICCV}, 2023.

\bibitem[Yin et~al.(2024)Yin, Liu, Xiao, Cohen-Or, Huang, and Chen]{sai3d}
Yingda Yin, Yuzheng Liu, Yang Xiao, Daniel Cohen-Or, Jingwei Huang, and Baoquan Chen.
\newblock Sai3d: Segment any instance in 3d scenes.
\newblock In \emph{CVPR}, 2024.

\bibitem[Zhang et~al.(2023{\natexlab{a}})Zhang, Han, Qiao, Kim, Bae, Lee, and Hong]{mobile_sam}
Chaoning Zhang, Dongshen Han, Yu Qiao, Jung~Uk Kim, Sung-Ho Bae, Seungkyu Lee, and Choong~Seon Hong.
\newblock Faster segment anything: Towards lightweight sam for mobile applications.
\newblock \emph{arXiv preprint arXiv:2306.14289}, 2023{\natexlab{a}}.

\bibitem[Zhang et~al.(2023{\natexlab{b}})Zhang, Liang, Yang, Zou, Ye, Liu, and Bai]{zhang2023sam3d}
Dingyuan Zhang, Dingkang Liang, Hongcheng Yang, Zhikang Zou, Xiaoqing Ye, Zhe Liu, and Xiang Bai.
\newblock Sam3d: Zero-shot 3d object detection via segment anything model.
\newblock \emph{arXiv preprint arXiv:2306.02245}, 2023{\natexlab{b}}.

\bibitem[Zhao et~al.(2021)Zhao, Jiang, Jia, Torr, and Koltun]{pointtrans}
Hengshuang Zhao, Li Jiang, Jiaya Jia, Philip Torr, and Vladlen Koltun.
\newblock Point transformer.
\newblock In \emph{ICCV}, 2021.

\bibitem[Zhao et~al.(2020)Zhao, Chua, and Lee]{zhao2020sess}
Na Zhao, Tat-Seng Chua, and Gim~Hee Lee.
\newblock Sess: Self-ensembling semi-supervised 3d object detection.
\newblock In \emph{CVPR}, 2020.

\bibitem[Zou et~al.(2023{\natexlab{a}})Zou, Dou, Yang, Gan, Li, Li, Dai, Behl, Wang, Yuan, Peng, Wang, Lee, and Gao]{zou2023xdecoder}
Xueyan Zou, Zi-Yi Dou, Jianwei Yang, Zhe Gan, Linjie Li, Chunyuan Li, Xiyang Dai, Harkirat Behl, Jianfeng Wang, Lu Yuan, Nanyun Peng, Lijuan Wang, Yong~Jae Lee, and Jianfeng Gao.
\newblock Generalized decoding for pixel, image, and language.
\newblock In \emph{CVPR}, 2023{\natexlab{a}}.

\bibitem[Zou et~al.(2023{\natexlab{b}})Zou, Yang, Zhang, Li, Li, Gao, and Lee]{zou2023seem}
Xueyan Zou, Jianwei Yang, Hao Zhang, Feng Li, Linjie Li, Jianfeng Gao, and Yong~Jae Lee.
\newblock Segment everything everywhere all at once.
\newblock In \emph{NeurIPS}, 2023{\natexlab{b}}.

\end{thebibliography}
}
\clearpage
\setcounter{figure}{0}
\setcounter{table}{0}

\begin{appendices}
\startcontents[supple]

{
    \hypersetup{linkcolor=black}
    \printcontents[supple]{}{1}{}
}
\renewcommand{\thefootnote}{\fnsymbol{footnote}}
\renewcommand{\thesection}{\Alph{section}}%
\renewcommand\thetable{\Roman{table}}
\renewcommand\thefigure{\Roman{figure}}

\section{More Qualitative Results }\label{sec:more_results}

\begin{figure*}[htbp]
    \centering
    \captionsetup{type=figure}
    \includegraphics[width=0.8\textwidth]{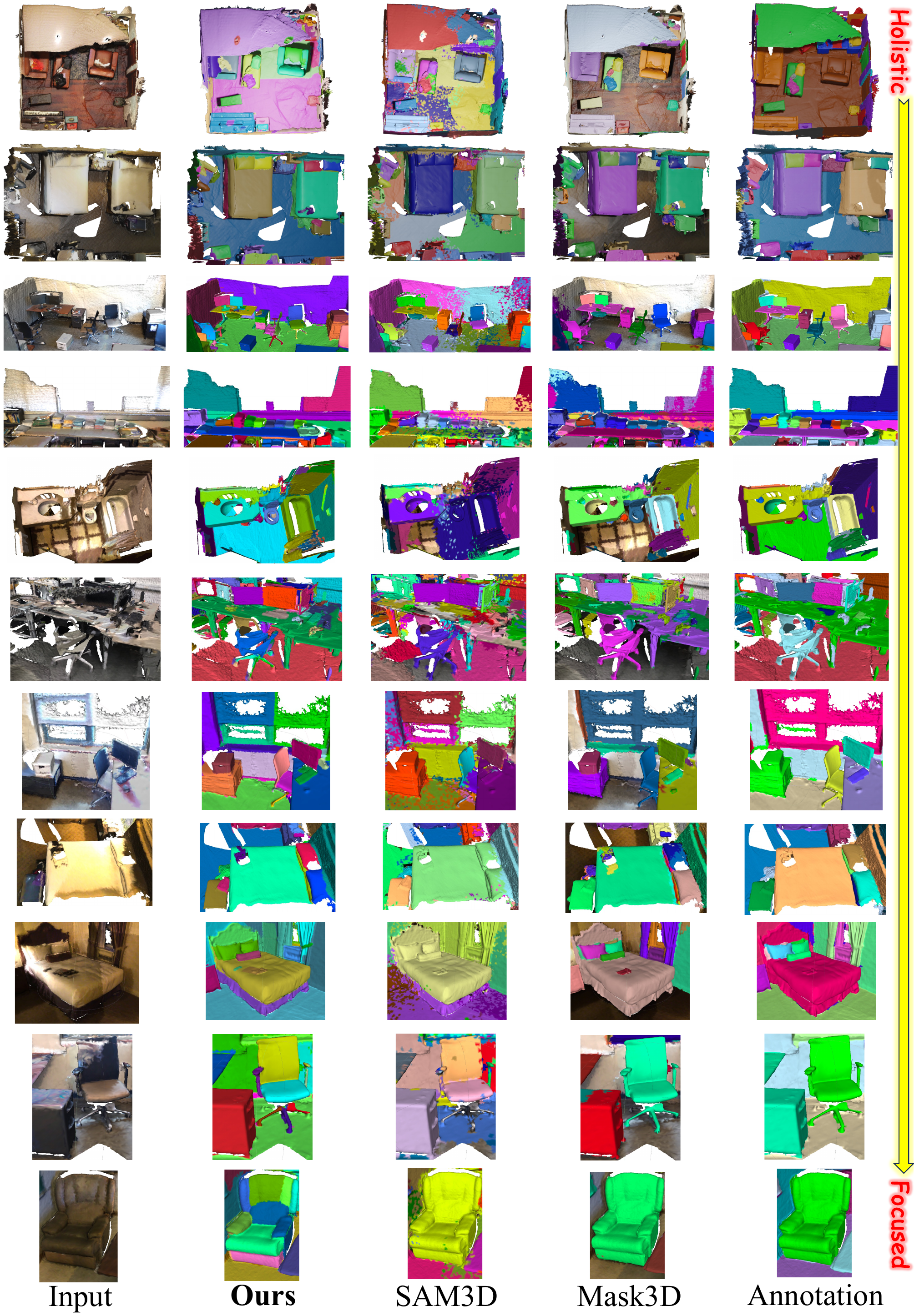}
    \captionof{figure}{The qualitative comparison of \textbf{our method, SAM3D~\cite{sam3d}, Mask3D~\cite{mask3d} and ScanNet200's annotations \cite{scannet200}}, across various scenes in the \textbf{ScanNet200 validation set}, from holistic to focused view. Note that Mask3D does not treat floor and wall as instances, resulting in the absence of these two labels in its results.
    } \label{fig:qualitative_result_supple}
\end{figure*}%

\subsection{On ScanNet200}
Following the main paper, \cref{fig:qualitative_result_supple} presents more qualitative comparisons on the ScanNet200 validation set, where our method is compared with SAM3D \cite{sam3d}, Mask3D \cite{mask3d}, and the original annotations of ScanNet200 \cite{scannet200}.
Note that Mask3D does not treat floor and wall as instances, resulting in the absence of these two labels in its results.

Consequently, our method consistently achieves remarkable 3D scene segmentation results across diverse scenes and objects, from holistic views to focused perspectives.
Notably, our approach significantly outperforms SAM3D in terms of segmentation quality and diversity. 
When compared to Mask3D (\textit{trained and evaluated both on ScanNet200}), our method demonstrates competitive or superior segmentation quality and diversity.
Importantly, our results not only match the quality of human annotations but also exhibit greater diversity in many cases.


\subsection{On ScanNet200-Fine50}
\begin{figure*}[t]
    \centering
    \captionsetup{type=figure}
    \includegraphics[width=0.6\linewidth]{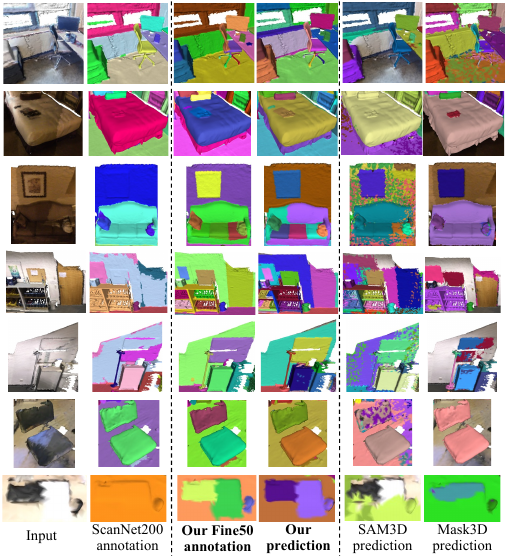}
    \captionof{figure}{The qualitative comparison among \textbf{the predictions from our method, SAM3D \cite{sam3d}, and Mask3D \cite{mask3d}, as well as the original annotations from ScanNet200 \cite{scannet200} and the fine-grained annotations from our ScanNet200-Fine50}, across diverse scenes from \textbf{focused views}. Note that Mask3D does not treat floor and wall as instances, resulting in the absence of these two labels in its results.}
    \label{fig:fine50_result}
\end{figure*}

We have introduced a fine-grained test set called ScanNet200-Fine50. In \cref{fig:fine50_result}, we provide more qualitative comparisons among the predictions from our method, SAM3D \cite{sam3d}, and Mask3D \cite{mask3d}, as well as the original annotations from ScanNet200 \cite{scannet200} and the fine-grained annotations from our ScanNet200-Fine50. Here, we mainly show \textit{focused} views, aiming to highlight the segmentation performance specifically for fine-grained instances.

In comparison to the initial annotations in ScanNet200, our ScanNet200-Fine50 test set offers significantly more finely detailed annotations of high quality. Furthermore, our method's predictions exhibit better alignment with ScanNet200-Fine50 when compared to SAM3D and Mask3D, demonstrating the fine-grained segmentation capability of our approach.



\section{Results on Matterport3D}\label{sec:matterport_results}
As mentioned in the main paper, we also applied our method to the Matterport3D dataset \cite{matterport3D}.
In this dataset, the RGB frames exhibit \textit{larger view changes} compared to ScanNet \cite{scannet} and ScanNet++ \cite{scannetpp}, which presents additional challenges when performing segmentation solely on 2D frames. We follow \cite{openscene} to use undistorted images in the official Matterport3D repo $^{\ref{footnote:matterport_repo}}$ and test on 160 classes of the validation set.

\footnotetext[1]{\label{footnote:matterport_repo}\url{https://github.com/niessner/Matterport}}

\begin{table}[t]
\centering
\resizebox{0.6\linewidth}{!}{
\begin{tabular}{l | c | c | c}
\toprule
Model & $\mathbf{AP}$ & $\mathbf{AP_{50}}$ & $\mathbf{AP_{25}}$\\
\midrule
SAM3D \cite{sam3d} & 10.1 & 19.4 & 36.1\\
SAI3D \cite{sai3d} & 21.5 & 38.3 & 59.1\\
\textbf{Ours} & \textbf{24.3} & \textbf{42.1} & \textbf{65.4}\\
\bottomrule
\end{tabular}
}
\caption{The quantitative comparison on \textbf{Matterport3D} \cite{scannet200}.}\label{tab:matterport_result}
\end{table}

\begin{figure}[t]
    \centering
    \includegraphics[width=0.66\linewidth]{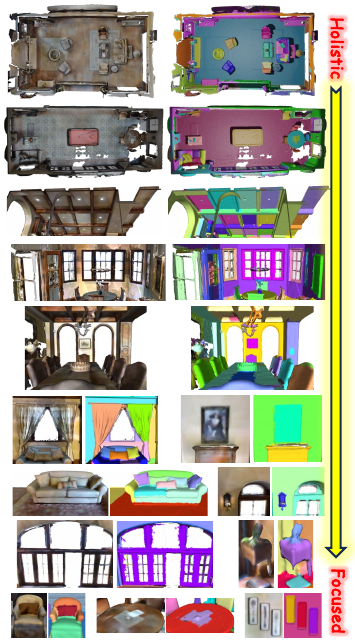}
    \captionof{figure}{The qualitative results of our method on \textbf{Matterport3D} \cite{matterport3D}, from holistic to focused view. The results are arranged in pairs where the left is the input and the right is our output.}
    \label{fig:matterport_result}
\end{figure}%

As shown in \cref{tab:matterport_result}, our method outperforms both SAM3D \cite{sam3d} and SAI3D \cite{sai3d} on Matterport3D dataset, which is further supported by the qualitative results in \cref{fig:matterport_result}.
This further proves the robustness of our method on novel 3D scenes.

\section{An Augmented 2D Propagation}\label{sec:new_sam-pt}
In \cref{sec:intro} and \cref{fig:key_idea} (c) of the main paper, we discuss achieving prompt consistency by using automatic-SAM \cite{kirillov2023sam} on an initial frame to generate pixel prompts which can be propagated to subsequent frames, similar to SAM-PT \cite{sam-pt} for video tracking. 
However, prompts generated on initial frames of 3D scenes cannot cover newly emerged instances in other frames, leading to the absence of segmentation for many instances.

In this section, we evaluate an alternative scheme. Instead of performing automatic-SAM only once on an initial frame, we check if any areas lack segmentation masks in a frame, indicating the presence of newly emerged objects. In such cases, we reapply automatic-SAM to make prompts cover these objects.

\begin{table}[t]
\centering
\resizebox{0.8\linewidth}{!}{
\begin{tabular}{l | cc | cc | cc}
\toprule
Model & $\mathbf{AP}$ & $\mathbf{AP_{50}}$ & $\mathbf{AP_{25}}$\\
\midrule
Augmented 2D propagation & 20.3 & 38.6 & 59.7 \\
\textbf{Ours} & \textbf{26.3} & \textbf{47.2} & \textbf{68.6}\\
\midrule
\textcolor{blue}{Ours+HQ-SAM~\cite{sam_hq}} & \textcolor{blue}{28.5} & \textcolor{blue}{47.9} & \textcolor{blue}{69.8}\\
Ours+Mobile-SAM~\cite{mobile_sam} & 20.9 & 40.8 & 61.3 \\
\bottomrule
\end{tabular}
}
\caption{The quantitative results on ScanNet200 \cite{scannet200}. “\textbf{Augmented 2D propagation}” is detailed in \cref{sec:new_sam-pt}. “+HQ.” and “+Mob.” respectively indicate \textbf{incorporating HQ-SAM \cite{sam_hq} and Mobile-SAM \cite{mobile_sam}} in our framework.}\label{tab:quant_ablation_supple}
\end{table}

\begin{figure}[t]
    \centering
    \includegraphics[width=0.7\linewidth]{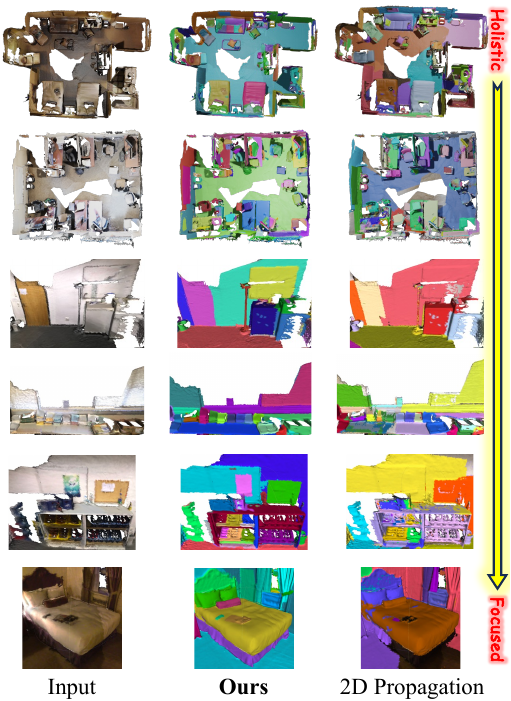}
    \captionof{figure}{The qualitative comparison between \textbf{our method and the augmented 2D propagation}, across various scenes in the ScanNet200 
    \cite{scannet200} validation set, from holistic to focused view.} 
    \label{fig:aug_pt_quali}
\end{figure}

As depicted in \cref{fig:aug_pt_quali}, although the augmented version of 2D propagation improves the completeness of 3D segmentation results, it still falls short in terms of both segmentation quality and diversity. Its deficiency is further highlighted by the comparison of mAP scores in \cref{tab:quant_ablation_supple}. 
The primary reason behind this inferior performance is that the augmented scheme only aligns pixel prompts within a \textit{limited} range, from the frame where automatic-SAM is applied to the next time reapplying it. 
Consequently, the mask consistency is restricted to a few frames. 
In contrast, our 3D prompts globally align pixel prompts across \textit{all} frames, resulting in comprehensive frame-consistent pixel prompts and 2D masks, as well as superior 3D segmentation results.

\section{More Ablation Studies}\label{sec:more_ablation}

\begin{table}[t]
\centering
\setlength{\tabcolsep}{3.0pt}
\resizebox{1.0\linewidth}{!}{
\begin{tabular}{c|cc|cccc|cccccc|cc}
\toprule
& \multicolumn{2}{c|}{\textbf{Module Impact}} 
& \multicolumn{4}{c|}{\textbf{Initial Prompts}} 
& \multicolumn{6}{c|}{\textbf{$\mathtt{\theta}_{retain}$}}
& \multicolumn{2}{c}{\textbf{Selection}} \\
 & \textcolor{orange}{w/o Sel.} & \textcolor{orange}{w/o Con.}& 1\% & 1.5\% & 2\% & \textcolor{orange}{5\%} & \textcolor{orange}{0.3} & 0.4 & 0.5 & 0.6 & 0.7 & \textcolor{orange}{0.8} & soft & top-k \\
 \midrule
Normal & 43.2 & 42.6 & 46.7 & 47.8 & \textbf{48.1} & 44.2 & 40.7 & 47.9 & \textbf{48.1} & 47.9 & 47.3 & 42.5 & 47.5 & 47.7\\
\textcolor{blue}{Small} & 25.6 & 25.1 & \textcolor{orange}{29.1} & 30.1 & \textbf{30.3} & 26.1 & 25.6 & \textcolor{orange}{29.2} & \textbf{30.3} & 30.1 & 29.9 & 25.7 & 30.0 & 29.8\\
\textcolor{blue}{Tiny} & 22.9 & 22.5 & \textcolor{orange}{24.3} & 25.3 & \textbf{25.6} & 23.7 & 23.2 & \textcolor{orange}{24.1} & \textbf{25.6} & 25.4 & 24.9 & 23.0 & 25.2 & 25.0\\
\bottomrule
\end{tabular}
}
\caption{The quantitative ablation studies \textbf{on our ScanNet200-Fine50} test set. We report $\mathbf{AP_{50}}$ across different mask sizes of our GT annotations (Normal, Small, Tiny). “w/o Sel.” and “w/o Con.” respectively denote discarding prompt selection and consolidation. We also evaluate our method using different ratios (1\%, 1.5\%, 2\%, 5\%) of input points as our initial prompts. $\mathtt{\theta}_{retain}$ is the threshold in prompt selection. “soft” and “top-k” are two voting schemes used during prompt selection.}\label{tab:quant_ablation_fine50}
\end{table}

\subsection{The Ablation Results on ScanNet200-Fine50}
Following the main paper, we conduct similar ablation studies on our ScanNet200-Fine50 test set and report the result in \cref{tab:quant_ablation_fine50}.
Similar to the conclusions in the main paper, removing prompt selection or consolidation leads to a performance drop across different mask sizes.
Besides, using 5\% initial prompts or setting $\mathtt{\theta}_{retain}=0.3, 0.8$ results in worse performance.

However, the fine-grained segmentation results (Small, Tiny) also \textit{slightly} degrade when using 1\% initial prompts or setting $\mathtt{\theta}_{retain}=0.4$. One possible reason is that when selecting fewer prompts, prompts may have a lower probability of being \textit{accurately scattered} onto fine-grained instances, as this kind of instance only occupies a small area. In this scenario, prompts may tend to be located on large-sized instances, resulting in the absence of segmentation for fine-grained instances.

\subsection{Frame Gaps}
\begin{figure}[t]
    \centering 
    \captionsetup{type=figure}
    \includegraphics[width=0.5\linewidth]{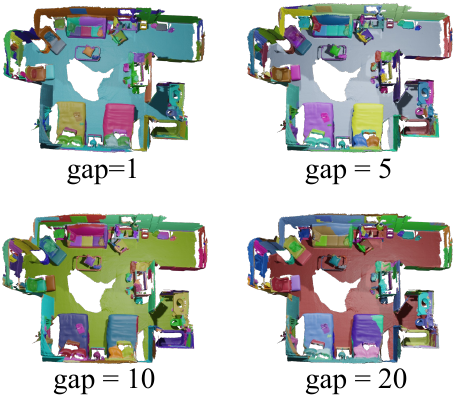}
    \captionof{figure}{The qualitative results of using \textbf{different frame gaps} in our method.
    } \label{fig:qualitative_frame_gap}
\end{figure}

\begin{figure}[t]
    \centering
    \captionsetup{type=figure}
    \includegraphics[width=0.6\linewidth]{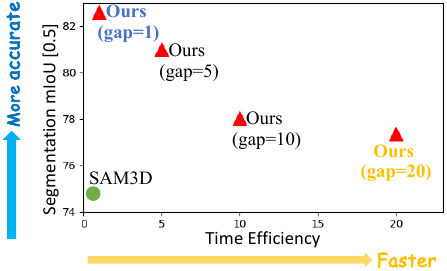}
    \captionof{figure}{The quantitative results ($AP_{50}$ on ScanNet200 and time costs) of using \textbf{different frame gaps} in our method. We also compare them against SAM3D \cite{sam3d}.
    } \label{fig:quantitative_frame_gap}
\end{figure}

In the context of performing SAM \cite{kirillov2023sam} on 2D image frames, an alternative approach is to skip frames with a certain gap. \cref{fig:qualitative_frame_gap} illustrates the qualitative results obtained by skipping frames with different gap numbers.
\cref{fig:quantitative_frame_gap} depicts the quantitative results on the ScanNet200 validation set \cite{scannet200}, considering both segmentation $AP_{50}$ and time cost.

The results indicate that the segmentation accuracy remains satisfactory with a gap of 5, while there is a degradation when using a gap of 10 or 20. 
Besides, according to \cref{fig:qualitative_frame_gap}, our framework stably maintains good segmentation diversity across different gap settings.
It's also worth mentioning that our method consistently outperforms SAM3D \cite{sam3d} in view of both segmentation accuracy and efficiency under different frame gaps, as indicated in \cref{fig:quantitative_frame_gap}.
To summarize, increasing the number of frame gaps reduces the number of frames on which SAM is applied, resulting in lower time costs. Therefore, one can choose a suitable frame gap that strikes a balance between segmentation quality and efficiency.

\section{Integrating Variants of SAM}\label{sec:sam_based_models}

\begin{figure}[t]
    \centering
    \captionsetup{type=figure}
    \includegraphics[width=0.8\linewidth]{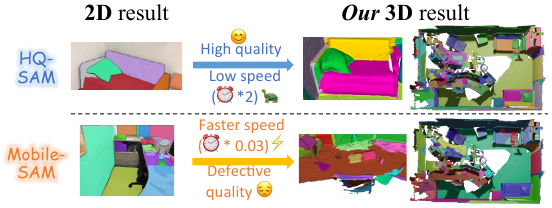}
    \captionof{figure}{The qualitative result of \textbf{incorporating HQ-SAM \cite{sam_hq} and Mobile-SAM \cite{mobile_sam}} into our framework.
    } \label{fig:hq_mobile_sam}
\end{figure}

Our method can serve as a general framework to integrate the models based on SAM \cite{kirillov2023sam}.
We first integrate HQ-SAM \cite{sam_hq} and Mobile-SAM \cite{mobile_sam} into our framework. As depicted in \cref{fig:hq_mobile_sam} and \cref{tab:quant_ablation_supple}, their impact on 2D images seamlessly translates to the improved performance in our method.
These experimental results not only validate the \textit{versatility} of our method but also support a fundamental concept driving our motivation, which is to directly transfer the power of SAM or its variants into 3D without sophisticated training.
This insight highlights the importance of considering the holistic system rather than solely focusing on pure 3D data representation in future research endeavors.

\begin{figure}[t]
    \centering
    \includegraphics[width=0.66\linewidth]{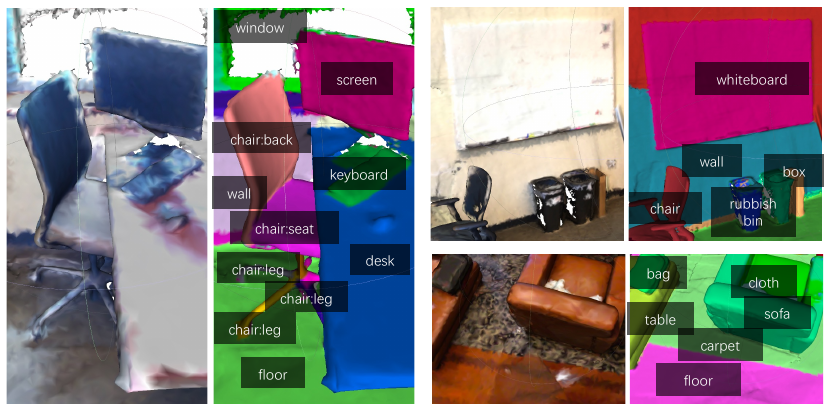}
   \caption{The qualitative result of \textbf{integrating Semantic-SAM \cite{li2023semanticsam}} into our framework.} \label{fig:semantic_sam}
\end{figure}

In addition, Semantic-SAM \cite{li2023semanticsam} recently brought semantic awareness to SAM. By integrating Semantic-SAM into our framework, we can also achieve 3D semantic understanding. In this scheme, each point prediction from SAM can be enhanced with the semantic label that is assigned by Semantic-SAM across 2D frames the most number of times. \cref{fig:semantic_sam} shows some pilot results.

\section{Implementation Details}\label{sec:implementation}

\paragraph{Details of running SAM.}

When performing SAM \cite{kirillov2023sam} segmentation on all 2D frames, we use the PyTorch \cite{pytorch} code provided in the original SAM repository~$^{\ref{footnote:sam_code}}$.
Following this code, we begin by converting all the projected pixel coordinates into batched torch tensors represented as an array (BxNx2), where each element corresponds to the (u, v) coordinates in pixels. Next, we pass these tensors to the SAM predictor to process all the pixel prompts in parallel. For this process, we consider all pixel prompts as foreground pixels and assign them a label of 1.
We employ the ViT-H SAM \cite{kirillov2023sam} model, which is the default public model of SAM. We resize each RGB image frame to a resolution of 240$\times$320. Through experimentation, we found that this resolution is adequate for SAM to deliver satisfactory results. Therefore, we choose this resolution to optimize efficiency in our pipeline.

\footnotetext[2]{\label{footnote:sam_code}\url{https://github.com/facebookresearch/segment-anything}}

\paragraph{Details of View-Guided Prompt Selection.}
As mentioned in \cref{sec:prompt_filter} of the main paper, the first step of View-Guided Prompt Selection is to select prompts on each individual frame.
In detail, this selection process on individual frames involved three steps which are based on the strategy proposed in automatic-SAM \cite{kirillov2023sam}.
Firstly, to \textit{eliminate overlapped} masks, we employed standard greedy box-based non-maximal suppression (NMS) and cut the masks with a box IoU smaller than 80.0. Secondly, we retained only \textit{confident} masks by applying a threshold of 70.0 to the model's predicted IoU scores. Finally, we focused on \textit{stable} masks by comparing pairs of binary masks derived from the same soft mask. We kept the prediction (\ie, binary mask resulting from thresholding logits at 0) if the IoU between its pair of -1 and +1 thresholded masks was 60.0 or higher.


\paragraph{Details of building ScanNet200-Fine50.}
To build our ScanNet200-Fine50 test set, we handpicked 50 scenes from the ScanNet200 \cite{scannet200} validation set. These selected scenes predominantly feature a higher number of fine-grained instances that lack mask annotations, such as multiple small instances on tables.
Subsequently, we engaged the expertise of five experienced 3D data annotators, assigning each of them 10 scenes for annotation.
Throughout the annotation process, they were instructed to meticulously examine each instance and provide annotations with the utmost level of detail possible. For instance, their annotations encompass not only each small instance on a table but also different removable parts of a chair. At present, our fine-grained annotations are agnostic to specific categories and do not include explicit category labels. Furthermore, the annotators will cross-check the initial annotations provided by their peers and offer feedback through an online communication system. This process ensures the meticulous and high-quality annotation of the data.

\section{A Supplementary Evaluation Scheme}\label{sec:grouping}

\begin{figure}[t]
\centering
    \centering
    \captionsetup{type=figure}
    \includegraphics[width=0.5\linewidth]{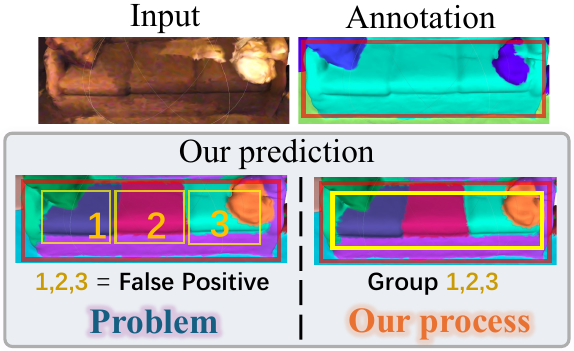}
    \captionof{figure}{Evaluation of the fine-grained predictions that lack GT annotations. Predictions 1,2,3 (sofa cushions) are \textbf{visually good}. However, if we compare them with the whole annotated sofa, they will all be \textbf{False} Positive due to small IoU. Thus, we group them for better evaluation.} \label{fig:mAP}
\end{figure}

As mentioned in the main paper, different from previous zero-shot or fully-supervised methods, our approach preserves the zero-shot power of SAM, \textit{often} segmenting \textit{fine-grained} instances that \textit{lack} corresponding accurate Ground Truth (GT) annotations (as in \cref{fig:qualitative_result_supple}).
Consequently, as illustrated in \cref{fig:mAP} (“Problem”), if we directly compare our predictions with GT annotations during mAP calculation, our successfully-segmented fine-grained instances will be counted as False Positive, which hurts the accurate evaluation.
A similar problem occurs in the SAM paper \cite{kirillov2023sam} (Tab.~5), where evaluating SAM on classical coarsely-annotated datasets \cite{coco,gupta2019lvis} leads to inferior mAP results compared to the fully-supervised ViTDet \cite{vitdet}. However, SAM outperforms ViTDet according to the user study.

To handle this issue, we add a grouping process. We begin by selecting an annotated instance $\mathbf{g}$ from the validation data. We then traverse our segmentation outputs to identify all predictions $\{O_m|i=m,...,M\}$ that \textit{belonging} to $\mathbf{g}$ by checking if the most area ($>80\%$) of $O_m$ is included by $\mathbf{g}$. 
We group all such predictions $O_m$ into a single prediction which is then compared with $\mathbf{g}$ to decide whether it is a True Positive, as shown in \cref{fig:mAP} (“Our process”). This process is repeated for all annotated instances.

\begin{table}[t]
\centering
\resizebox{1.0\linewidth}{!}{
\begin{tabular}{lcc|cc|cc}
\toprule
 & \multicolumn{2}{c|}{$\mathbf{AP}$} & 
\multicolumn{2}{c|}{$\mathbf{AP_{50}}$} &
\multicolumn{2}{c}{$\mathbf{AP_{25}}$}\\
Method 
 & w/o group & \textbf{w/ group} 
 & w/o group & \textbf{w/ group}  
 & w/o group & \textbf{w/ group} \\
\midrule
Mask3D \cite{mask3d} & 53.3 & 54.0 (\textcolor{orange}{0.7$\uparrow$})
                     & 71.9 & 72.7 (\textcolor{orange}{0.8$\uparrow$})
                     & 81.6 & 82.2 (\textcolor{orange}{0.6$\uparrow$})\\
\textbf{Ours}        & 26.3 & 33.7 (\textcolor{blue}{7.4$\uparrow$}))
                     & 47.2 & 54.3 (\textcolor{blue}{7.1$\uparrow$})
                     & 68.6 & 77.2 (\textcolor{blue}{8.6$\uparrow$})\\
\bottomrule
\end{tabular}
}
\caption{Quantitative comparison with Mask3D (pretrained on ScanNet) on ScanNet200. “w/o group” and “w/ group” denote scores \textbf{without and with our grouping process}. While grouping operation \textcolor{blue}{improves} our scores, it has \textcolor{orange}{\textit{minimal} impact} on Mask3D which can always find GT annotations.}
\label{tab:grouping_result}
\end{table}

We apply the proposed grouping process when calculating the AP scores of Mask3D \cite{mask3d} and our method. The results on the ScanNet200 validation set are listed in \cref{tab:grouping_result}. First, it is evident that our grouping method has a \textit{minimal} impact on Mask3D.
This is primarily because Mask3D is pretrained on ScanNet200 training data, which possesses similar or even finer granularity compared to the test data. Consequently, most of Mask3D’s predictions successfully match the corresponding GT annotations in the validation set (as shown in \cref{fig:qualitative_result_supple}). In contrast, for our SAM-powered model, it is common for predictions to lack annotations, so the grouping operation leads to improved AP scores. Notably, our $\mathbf{AP_{25}}$ with grouping is even comparable with fully-supervised Mask3D. This experiment indicates that our grouping scheme not only mitigates the issue of evaluating predictions without annotations but also has minimal impact on evaluating the predictions with matched annotations, showcasing the fairness and rationality of our grouping process.

\stopcontents[supple]

\end{appendices}

\end{document}